\documentclass[10pt,twocolumn,letterpaper]{article}

\usepackage{cvm}
\usepackage{times}
\usepackage{epsfig}
\usepackage{graphicx}
\usepackage{amsmath}
\usepackage{amssymb}

\usepackage{subcaption}
\usepackage{mdframed}
\usepackage{color}
\usepackage{booktabs}
\usepackage{multirow, multicol}
\usepackage{xspace}
\usepackage[export]{adjustbox}
\usepackage{float}
\usepackage{makecell}

\makeatletter
\@namedef{ver@everyshi.sty}{}
\makeatother
\usepackage{tikz}
\usetikzlibrary{spy, calc}

\usepackage[pagebackref=true,breaklinks=true,letterpaper=true,colorlinks,bookmarks=false]{hyperref}

\newcommand{\palenerf}{PaletteNeRF\xspace}
\newlength{\imgw}
\usepackage{amsmath,amsfonts}

\newcommand{\vect}[1]{\mathbf{#1}}

\newcommand{\vc}{\vect{c}}
\newcommand{\vd}{\vect{d}}

\newcommand{\vo}{\vect{o}}
\newcommand{\vp}{\vect{p}}

\newcommand{\vr}{\vect{r}}

\newcommand{\vvv}{\vect{v}}
\newcommand{\vw}{\vect{w}}
\newcommand{\vx}{\vect{x}}

\newcommand{\MI}{\boldsymbol{I}}

\newcommand{\MV}{\boldsymbol{V}}
\newcommand{\MW}{\boldsymbol{W}}

\definecolor{dkyellow}{rgb}{0.5,0.5,0}
\definecolor{dkred}{rgb}{0.8,0,0}

\newcommand{\keywords}[1]{{\bf \emph{Keywords: #1}}}

\cvmfinalcopy 

\def\cvmPaperID{114} 

\ifcvmfinal\fi
\begin{document}

\title{\palenerf: Palette-based Color Editing for NeRFs}

\author{Qiling Wu\\
Tsinghua University\\
Qinghua Yuan, Haidian, Beijing\\
{\tt\small wql19@mails.tsinghua.edu.cn}
\and
Jianchao Tan\\
Kuaishou Technology\\
No.6 Shangdi West Road, Haidian, Beijing\\
{\tt\small tanjianchaoustc@gmail.com}
\and
Kun Xu\\
Tsinghua University\\
Qinghua Yuan, Haidian, Beijing\\
{\tt\small xukun@tsinghua.edu.cn}
}

\maketitle


\begin{abstract}
Neural Radiance Field (NeRF) is a powerful tool to faithfully generate novel views for scenes with only sparse captured images. Despite its strong capability for representing 3D scenes and their appearance, its editing ability is very limited.
In this paper, we propose a simple but effective extension of vanilla NeRF, named \emph{\palenerf}, to enable efficient color editing on NeRF-represented scenes. Motivated by recent palette-based image decomposition works, we approximate each pixel color as a sum of palette colors modulated by additive weights. Instead of predicting pixel colors as in vanilla NeRFs, our method predicts additive weights. The underlying NeRF backbone could also be replaced with more recent NeRF models such as KiloNeRF to achieve real-time editing. Experimental results demonstrate that our method achieves efficient, view-consistent, and artifact-free color editing on a wide range of NeRF-represented scenes. 
\end{abstract}


\section{Introduction}


Neural Radiance Field (NeRF)~\cite{mildenhall_nerf_2020} is a powerful tool for image-based modeling and rendering. With only a sparse set of captured images, it can faithfully generate rendering results from novel views. The core of NeRF is a neural volumetric representation of the scene using a multi-layer perceptron network. Due to its high effectiveness, it has attracted a wide range of attention from the community, with a bunch of follow-up works and applications since its introduction in 2020. However, due to the black-box nature of neural representations, all information, including geometries, materials, and light transports, are tightly baked into NeRFs, which are hard to be interpreted and edited. 

Recently, some methods have been proposed to enable color editing of NeRFs~\cite{liu_editnerf_2021, wang_clip-nerf_2022}.  Given a few user-provided color scribbles, EditingNeRF~\cite{liu_editnerf_2021} propagates the user edits to the whole data to achieve color editing and shape modification. However, this work is too demanding for datasets, which require many instances from the same category for training, limiting its practical usage. CLIP-NeRF~\cite{wang_clip-nerf_2022} uses embeddings of CLIP \cite{radford_clip_2021} to edit NeRFs. They finetune layers that influence color while freezing layers that influence density, to match the embedding of NeRF's output to that of the text editing prompt. However, it sometimes modifies undesired areas, leaving some artifacts in the results. PosterNeRF~\cite{tojo_poster-nerf_2022} 
gives an efficient way to extract palette from radiance fields, then utilizes posterization method~\cite{chao_posterchild_2021} to achieve real-time color editing. Despite the real-time performance, the editing results have artifacts like color banding and leaking as side effects of posterization.

\begin{figure}[tbp]
    \setlength\tabcolsep{1pt}
	\setlength{\imgw}{0.27\linewidth}
    \newcommand{\palew}{2mm}
	\begin{tabular}{@{}clllc@{}}
		\raisebox{0.5\imgw}{lego} & 
		\includegraphics[width=\imgw]{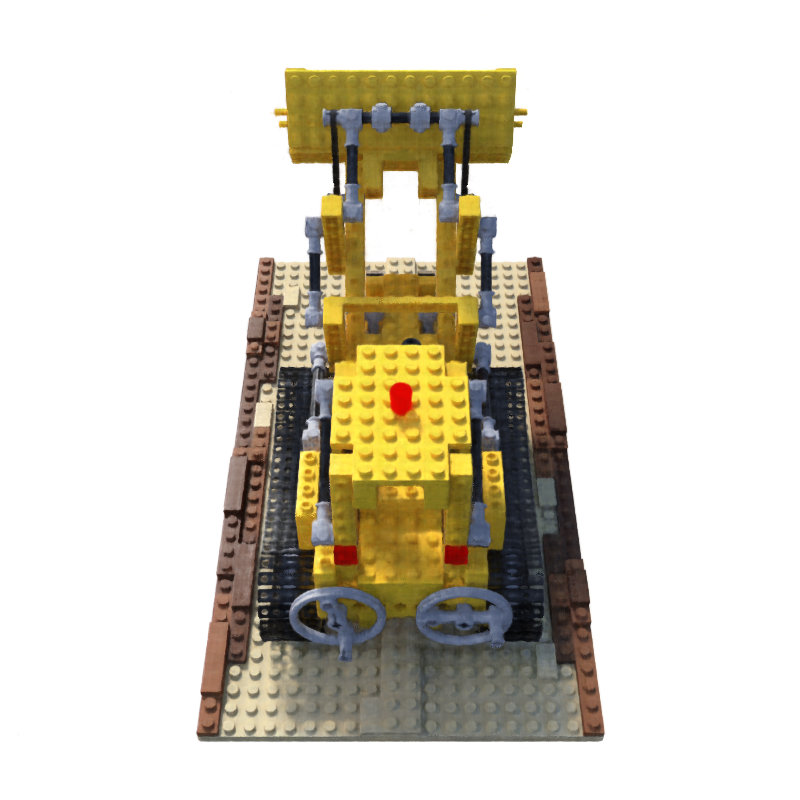} &
		\includegraphics[width=\imgw]{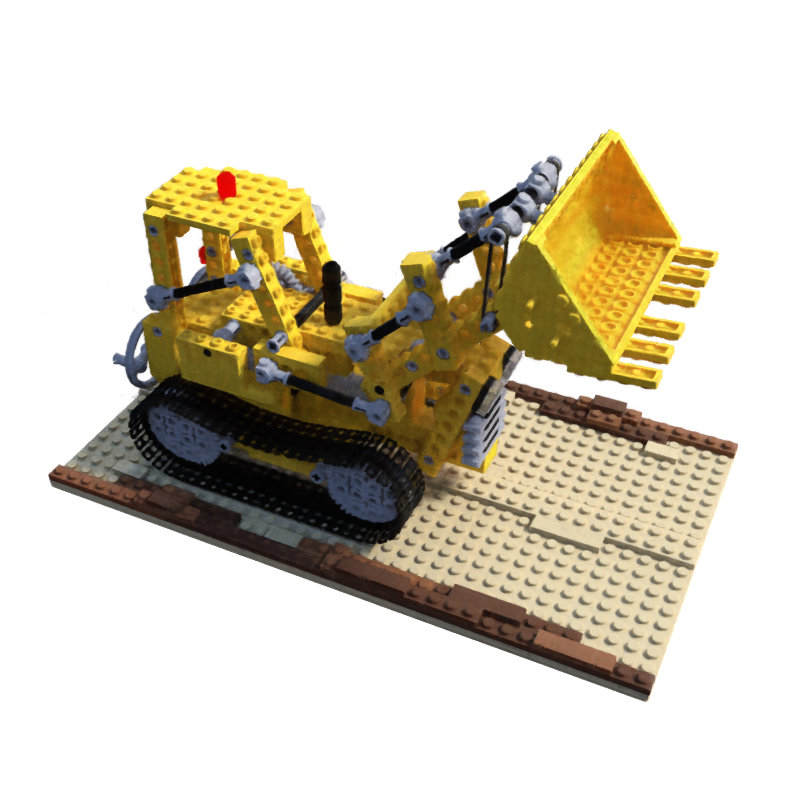} &
		\includegraphics[width=\imgw]{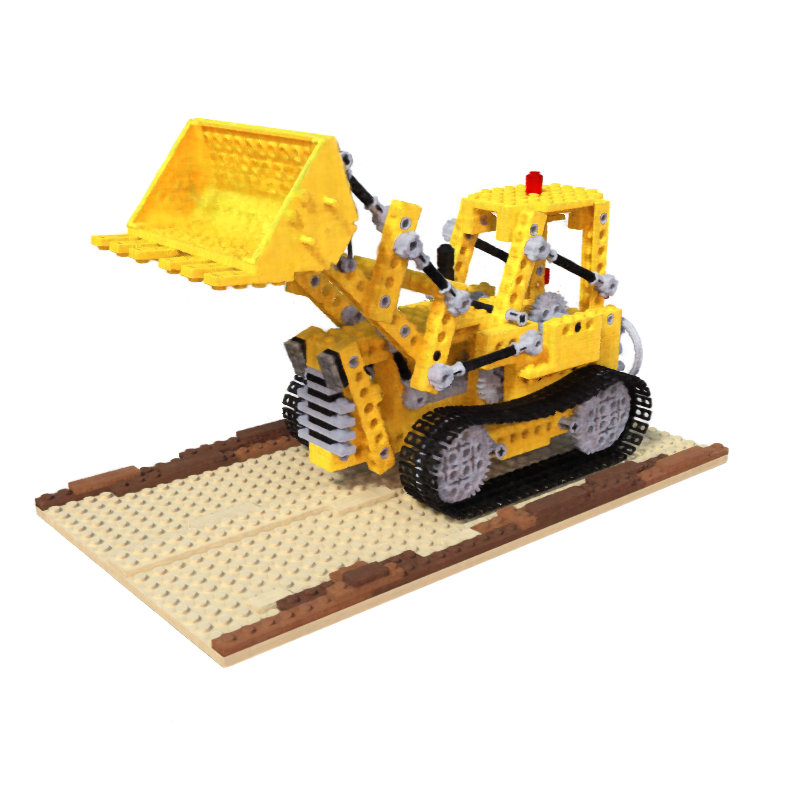} &
		\raisebox{0.5\imgw}{\rotatebox[origin=c]{-90}{Original}}
		\\
		\raisebox{3mm}{
			\begin{tikzpicture}[x=\palew,y=\palew]
				\node[] (before) at (0,0)
				{\adjustbox{frame}{\includegraphics[height=\palew, angle=-90]{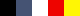}}};
				\node[] (after) at (2,0)
				{\adjustbox{frame}{\includegraphics[height=\palew, angle=-90]{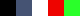}}};
				\draw[->, semithick] (0.5,-2) -- (1.5,-2);
			\end{tikzpicture}
		}
		&
		\includegraphics[width=\imgw]{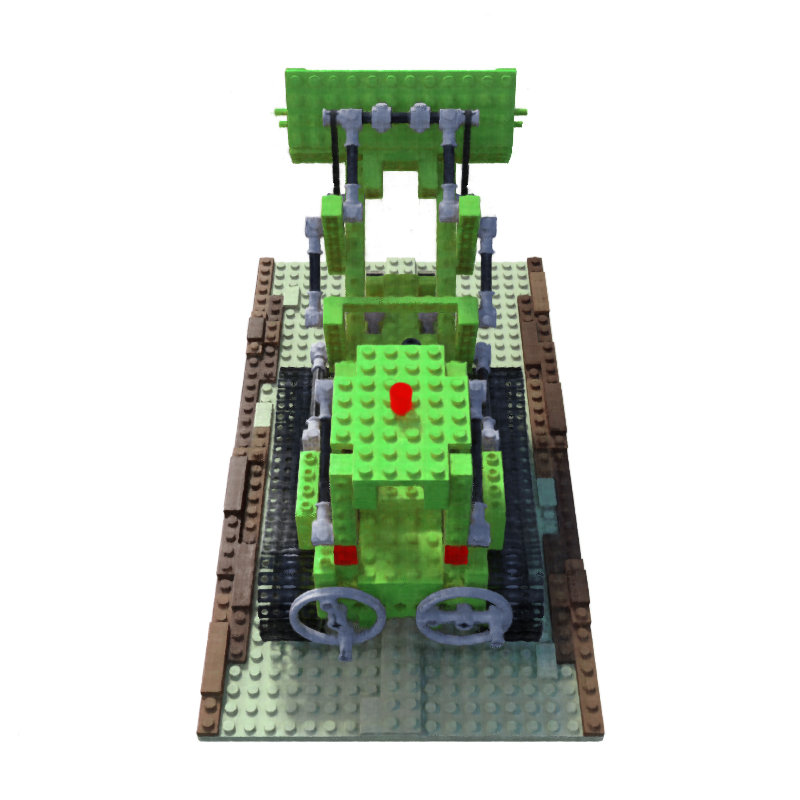} &
		\includegraphics[width=\imgw]{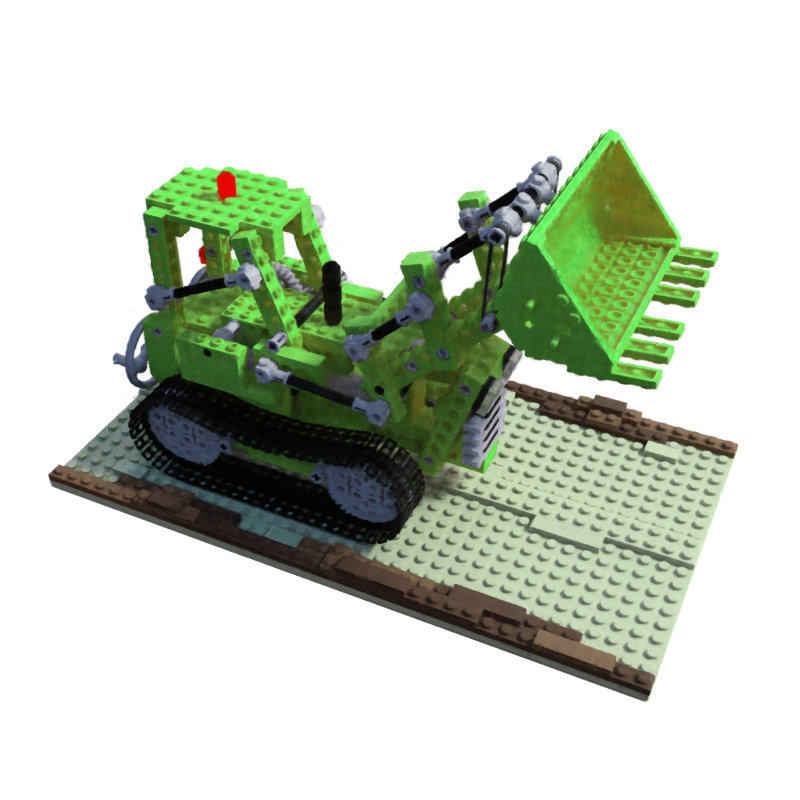} &
		\includegraphics[width=\imgw]{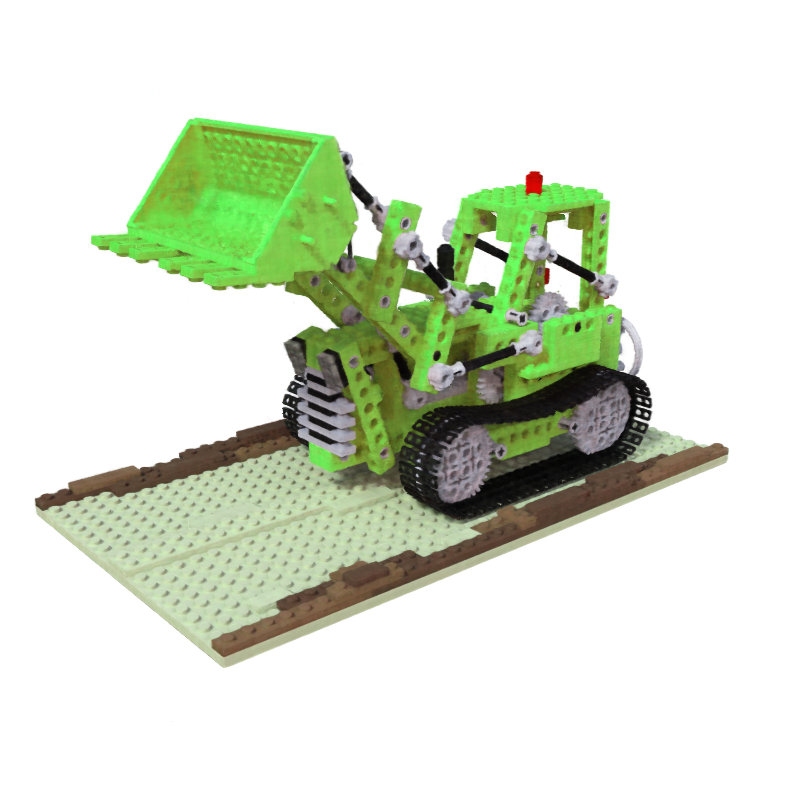} &
		\raisebox{0.5\imgw}{\rotatebox[origin=c]{-90}{Edited}}
	\end{tabular}
	\caption{Color editing results of scene \emph{lego} represented by Neural Radiance Field (NeRF). The origin and edited palettes are shown at the bottom-left corner. Users can modify the extracted palette to achieve intuitive, view-consistent, artifact-free editing of NeRFs.}
	\label{fig:teaser}
\end{figure}

In this paper, we propose \palenerf, an intuitive, view-consistent and artifact-free palette-based color editing method for NeRFs.
Inspired by palette-based image editing works, we approximate pixel colors as the sum of palette colors modulated by additive weights. Instead of predicting pixel colors as in vanilla NeRFs, we predict the high dimensional additive weights in our \palenerf. The NeRF-represented scenes can be recolored by adjusting the palette colors without retraining or modifying the \palenerf, as shown in Figure~\ref{fig:teaser}. The underlying NeRF backbone could also be replaced by more recent NeRF models such as KiloNeRF~\cite{reiser_kilonerf_2021} to achieve real-time editing.

\section{Related work}

\subsection{Neural Radiance Field}


Neural Radiance Field (NeRF)~\cite{mildenhall_nerf_2020} utilizes radiance field to model a 3D scene implicitly. Specifically, they use MLPs to infer volumetric density and radiance for certain points and view directions of a scene and follow the paradigm of volumetric rendering to compute the image pixel colors. Given a sparse set of captured images, NeRF can generate high-quality results for novel views. However, this framework needs to train a separate MLP for every scene. All information about the scene, including geometries, materials, and light transports, is baked into the neural representation. Hence, the vanilla NeRF does not allow changes in scene geometries, colors, and lighting.  
Various follow-ups of NeRFs have been proposed to address this limitation, i.e., to deal with deformable scenes~\cite{pumarola_d-nerf_2021, tretschk_non-rigid_2021, park_nerfies_2021, park_hypernerf_2021}, 
dynamic lighting~\cite{martin-nerf-inthewild_2021}, and scene composition~\cite{niemeyer_giraffe_2021, zhang_nerf++_2020, yu_compose_2022, yang_compose_2021, guo_osf_2020, ost_neural-scene-graph_2021}.

To improve NeRF's rendering speed, NSVF~\cite{liu_nsvf_2021} and KiloNeRF~\cite{reiser_kilonerf_2021} use empty space skipping and early ray termination. In addition, KiloNeRF divides the scene into small grids, and uses a small and efficient MLP in each grid to achieve further speedup. FastNeRF~\cite{garbin_fastnerf_2021} and PlenOctrees~\cite{yu_plenoctrees_2021} utilize function factorization and cache the MLP results to speedup rendering.
Some other methods aim at efficient training, e.g., Instant-NGP~\cite{muller_instantngp_2022} use hierarchical hash encoding to replace the costly MLP. 


As for editing, EditingNeRF~\cite{liu_editnerf_2021} takes a set of objects with similar shapes and colors, such as cars from Carla dataset~\cite{dosovitskiy_carla_2017}, then associate each instance with shape code and color code, which are fed into Conditional Radiance Fields (CRFs). The users provide a few color scribbles to indicate color changes and shape addition/removal of an object, which are propagated to specific regions of that object. However, it requires a set of shapes in the same class with different geometries and colors. CLIP-NeRF~\cite{wang_clip-nerf_2022} takes texts or exemplar images as edit prompts, which are fed into the multi-modal language model, i.e. CLIP, to obtain an embedding to serve as an offset for shape and color codes. The CRF process the modified codes to output edited results.
PosterNeRF~\cite{tojo_poster-nerf_2022} extracts palette efficiently from radiance fields. They sample RGB color points and use volumetric visibility to remove outliers, after which, a palette is extracted from the remaining points. Each pixel is approximated as a linear blending of at most 2 palette colors. Color editing is performed by adjusting the palette color. However, undesired artifacts can be easily perceived in the editing results.
In contrast, our work only requires a single scene to enable color editing. Our method provides a relatively simple user interface and achieves intuitive, artifact-free, and view-consistent color editing. Readers could refer to more details in the surveys on NeRFs~\cite{dellaert_survey-nerf_2021, tewari_survey-neu-render_2022}. 

\subsection{Palette based editing}
A palette is a concise representation of the color distributions of an image or video, which can be used to efficiently edit the image or video.
Several works have studied the human perception of the palette. These works construct various data fitting models according to human preference, and then regress a perceptual palette from input images~\cite{odonovan_color_2011, lin_modeling_2013, cao_mining_2017, feng_finding_2018}. 
Other works~\cite{chang_palette-based_2015, nguyen_group-theme_2017, zhang_palette-based_2017, aharoni-mack_pigment-based_2017} adopt clustering methods (e.g., k-means) over pixel colors, and use the cluster centers as palette colors. 
Another type of works~\cite{tan_decomposing_2017, tan_efficient_2019, wang_mvc_2019, delos_rgb_2019, grogan_image_2020, kim_automatic_2020} utilize geometric methods to generate palettes for images. Specifically, Tan et al.~\cite{tan_decomposing_2017, tan_efficient_2019} compute an RGB space convex hull that includes all colors of an image, followed by an iterative simplification of the convex hull until the vertex number is reduced to a predefined number or the reconstruction error reaches a predefined threshold. The vertices of the simplified convex hull are considered palette colors. 
Recently, Du et al.~\cite{du_video_2021} further extend this geometry-based method to time-lapse videos, generating time-varying palettes.
With the help of a palette, an image can be decomposed into multiple compositing layers, each of which contains spatially-variant per-pixel opacities or per-pixel mixing weights. 
Several methods~\cite{tan_decomposing_2015, tan_decomposing_2017} generate ordered translucent layers by assuming an alpha blending color composition mode. Other methods~\cite{aksoy_unmixing-based_2017, lin_layerbuilder_2017, tan_efficient_2019, wang_mvc_2019, Tan:2018:PPB} generate order-independent layers by assuming additive color composition mode. 
\section{Background}

\begin{figure*}[t]
    \centering
    \includegraphics[width=0.95\linewidth]{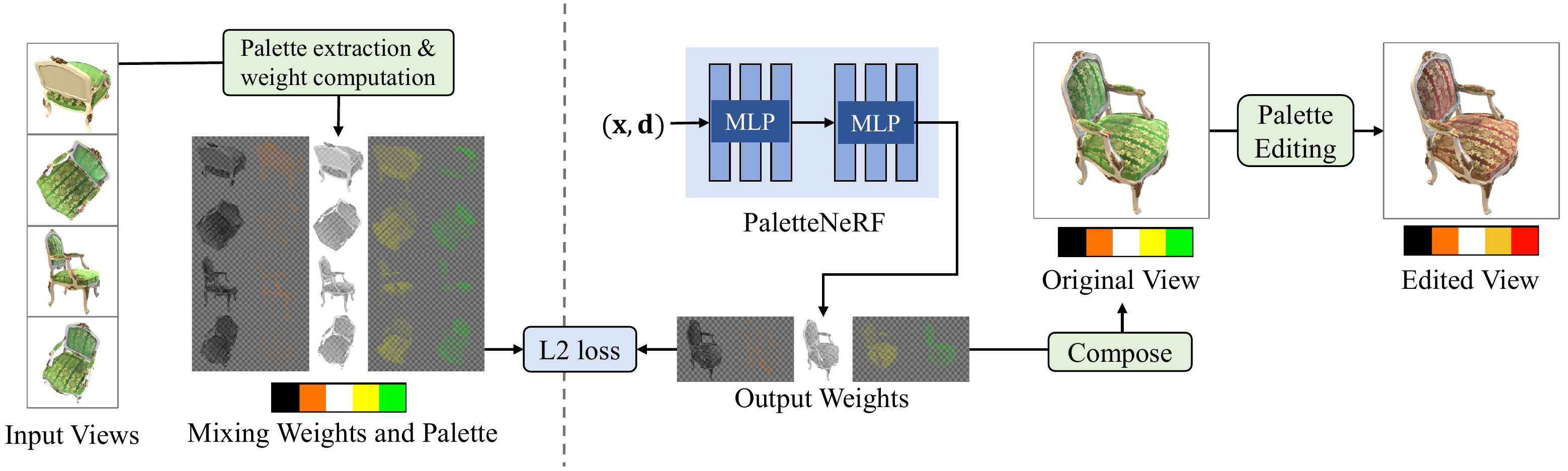}
    \caption{The pipeline of \palenerf. We first preprocess input views to obtain the palette and mixing weights. The weights are used as NeRF's fitting target. After training NeRF, we obtain output weights for a novel view, and modify the palette color to edit that view.}
    \label{fig:pipeline}
\end{figure*}

\subsection{Neural Radiance Field (NeRF)} 
Given a sparse set of captured images on a scene, NeRF~\cite{mildenhall_nerf_2020} can faithfully synthesize novel views for the scene. The core is a neural volumetric representation modeled with a multi-layer perceptron (MLP). The MLP 
$f$ predicts the RGB color and density observed at a 3D position $\vx$ from a view direction $\vd$:
\begin{equation} \label{equ:nerf_volume_representation}
    (\vc, \sigma) = f(\gamma(\vx), \gamma(\vd)), 
\end{equation}
where $\gamma(\cdot)$ denotes a positional encoding function. 


To render an image from a novel view, NeRF follows conventional ray marching techniques for volumetric rendering. For each image pixel, we first cast a ray $\vr$ from the viewpoint through the pixel to the scene. After that, several 3D points are sampled along the ray, and each sampled point is fed into the MLP to obtain its color and density. Finally, the colors of all sampled points are blended to obtain the rendered pixel color. 

A sparse set of captured images together with their camera parameters is sufficient for training a NeRF. A separate NeRF is required to be trained for each separate scene. During training, at each iteration, a batch of rays is randomly sampled from all (or a subset of) pixels. 
L2 differences between rendered pixel colors and ground truth pixel colors at given views are used as the loss function to be minimized. 


\subsection{Palette based image decomposition} 
Given an RGB image $\MI$, palette based image decomposition techniques~\cite{tan_decomposing_2017,tan_efficient_2019,wang_mvc_2019,du_video_2021} extract a small set of colors named \emph{palette} from an image, such that the color $\vc$ of each pixel $\vp$ could be approximately represented as a linear combination of palette colors: 
\begin{equation} \label{equ:palette_decomposition}
    \vc \approx \sum_{1\leq i \leq K} w_i^\vp \cdot \vvv_i, 
\end{equation}
where $K$ denotes palette size (i.e., number of palette colors) which usually varies from 4 to 10, depending on the color complexity of the input image. $\vvv_i$ denotes the $i$-th palette color, and $w_i^\vp$ denotes the additive weight of pixel $\vp$ with respect to palette color $\vvv_i$. The pixel-wise additive weights are required to satisfy two properties. First, they are non-negative: $w_i^\vp \geq 0 $ ($1\leq i \leq K$); second, they sum to one: $\sum_{i=1}^{K} w_i^\vp=1$. 

By denoting the palette as $\MV=[\vvv_1, \dots, \vvv_K]$ and the additive weight vector at each pixel $\vp$ as $\vw^\vp=[w_1^\vp, \dots, w_K^\vp]^T$, respectively, we could also rewrite the above formula in the vector-matrix form as:
\begin{equation} \label{equ:palette_decomposition2}
    \vc \approx \MV \cdot \vw^\vp. 
\end{equation}
By denoting the per-pixel additive weights ($\vw^\vp$) as a weight image $\MW$, i.e., having the same resolution as the input image $\MI$ while the channel size is changed from 3 to $K$, we could approximate pixel colors of the input image by:
\begin{equation} \label{equ:palette_decomposition3}
    \MI \approx \MV \cdot \MW. 
\end{equation}

There are various options of additive weights,  including as-sparse-as-possible weights~\cite{tan_decomposing_2017}, RGBXY barycentric weights~\cite{tan_efficient_2019}, and mean value coordinates (MVC)~\cite{wang_mvc_2019,ju_mvc_2005,floater_mvc_2005}. Since MVC weights could be efficiently computed and have been demonstrated to be smooth, sparse and effective~\cite{wang_mvc_2019,du_video_2021}, we use MVC weights in our paper.

Once the palette and the additive weight images are extracted, by simply adjusting the colors in the palette, the images can be instantly recolored through weighted linear interpolations from the palette colors (Eq.~\ref{equ:palette_decomposition}). This offers a more convenient editing interface than stroke-based methods, since those methods additionally user interaction steps, i.e., drawing strokes on the image. 

\section{Our Method} 


In this section, we introduce \palenerf, a concise modification to NeRF that makes scenes editable. We will first introduce our pipeline (Sec.~\ref{sec:overview}), then discuss the details of our network structure and training strategy (Sec.~\ref{sec:network_and_training}).

\begin{figure}[t]
    \includegraphics[width=1\linewidth]{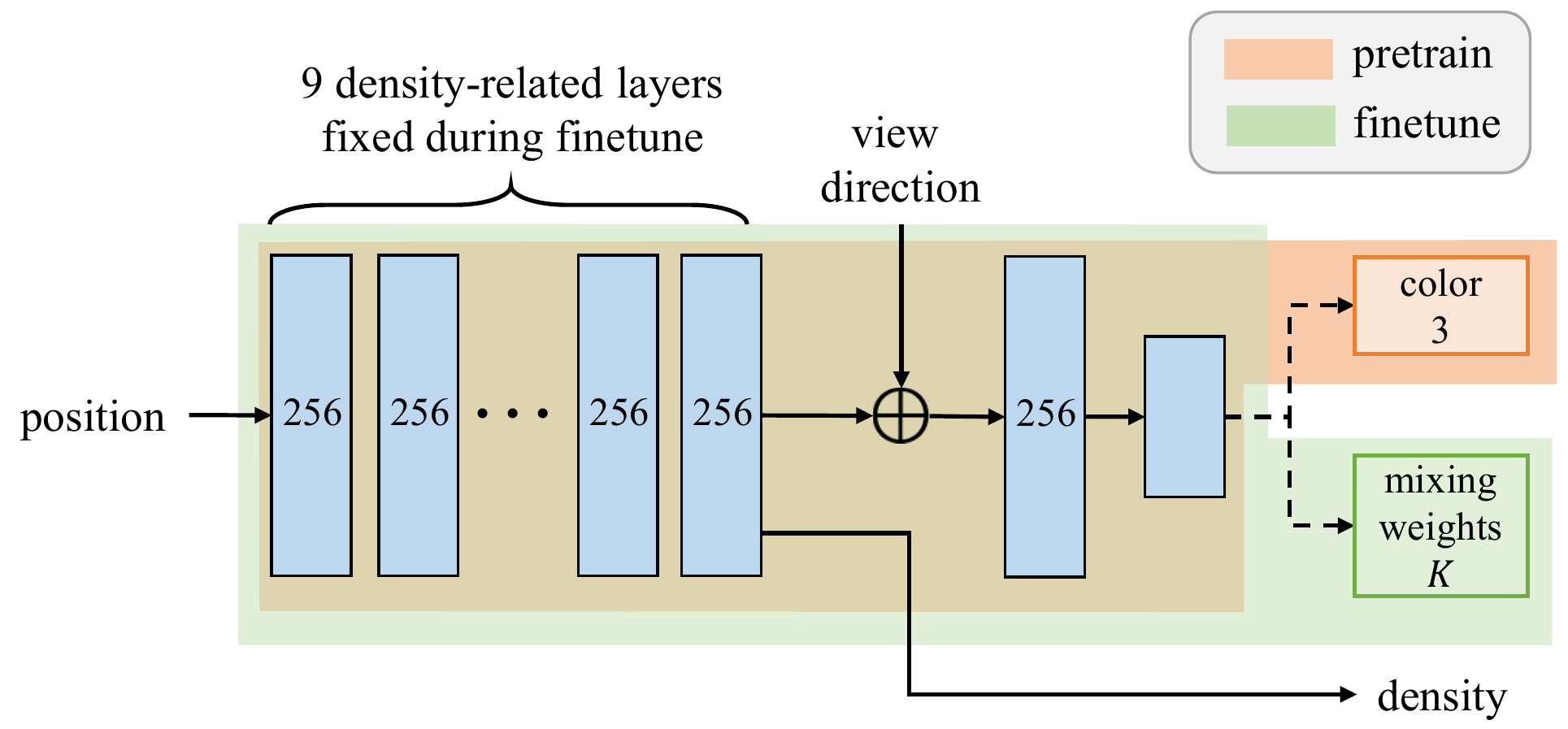}
    \caption{The network structure of our \palenerf. 
    The number inside each block denotes the dimension of the layer. 
    The top branch of the network predicts 3D colors which are used in the pretrain stage, while the bottom branch predicts $K$-channel additive mixing weights which are used in the finetune stage. 
    }
    \label{fig:fix-nerf}
\end{figure}

\subsection{Overview and Pipeline} \label{sec:overview}

Our goal is to provide an intuitive, efficient, and artifact-free color editing tool for NeRFs. 
Motivated by existing palette-based image decomposition works~\cite{tan_decomposing_2017,tan_efficient_2019,wang_mvc_2019,du_video_2021}, we find that if we approximate additive weights rather than pixel colors, the colors of the scene represented by NeRFs could be naturally edited. 

Specifically, instead of using NeRFs to model pixel colors which are essentially 3D (RGB) signals, we model higher-dimensional signals, e.g., the pixel-wise additive weights. Thanks to the high capacities of MLPs, the $K$-dimensional pixel-wise additive weights could also be well approximated. After that, we could intuitively edit the colors of the NeRF-represented scene by adjusting the palette colors. 
We refer to our modified NeRF as ~\emph{PaletteNeRF}.

Our overall pipeline is shown in Figure \ref{fig:pipeline}. The steps for utilizing \palenerf include:
\begin{itemize}
    \item \textbf{Data Preparation.} Given a set of sparse captured RGB images of a scene, we convert them to the same number of $K$-channel additive weight images and a shared palette $\MV$ with $K$ colors. This is done by stitching all images into a big image, and then directly applying the method in~\cite{wang_mvc_2019}. The pixel-wise additive weights are obtained using mean value coordinates. 
    \item \textbf{Modified MLP.} We feed the $K$-channel additive weight images to PaletteNeRF. Different from the vanilla NeRF which predicts 3-channel RGB colors, our PaletteNeRF predicts a $K$-channel vector $\vw$ indicating additive weights, i.e., its formulation is changed from $f(\vx, \vd) \rightarrow (\vc, \sigma)$ to  $f(\vx, \vd) \rightarrow (\vw, \sigma)$. 
    \item \textbf{Novel View Rendering.} To obtain the additive weight image $\MW_{\vo}$ from a novel view, we apply the same process as done in vanilla NeRF, i.e., through ray marching and blending of $K$-channel values. The RGB image from this novel view can be simply reconstructed through Eq.~\ref{equ:palette_decomposition} as: 
    \begin{equation}
        \MI_{\vo} = \MV \cdot \MW_{\vo}.
    \end{equation}
    \item \textbf{Color Editing.} Then, we can freely adjust the palette colors $\MV$ to achieve recoloring of the scene. During the color editing process, our \palenerf, which predicts additive weights, is not modified at all. This indicates that our editing process is decoupled with the \palenerf structure. Such decoupling has two advantages. First, the used NeRF backbone could be replaced with any newer, faster variants of NeRFs. In our experiments, we also demonstrate our method with a KiloNeRF backbone \cite{reiser_kilonerf_2021}, which allows color editing in real-time. Second, the consistency between different views is also naturally preserved, leading to artifact-free color editing with high fidelity. 
\end{itemize}


\subsection{Network Structure and Training} \label{sec:network_and_training}

As shown in Figure~\ref{fig:fix-nerf}, our network structure is almost the same as the vanilla NeRF, except for the last layer. We use a $128\times K$ fully connected (FC) layer, instead of a $128\times 3$ FC layer, as the last layer to output a $K$-channel vector. 

For training a \palenerf, we need to change the loss function from computing L2 differences between rendered and GT RGB images to computing L2 differences between rendered (predicted) and the GT $K$-channel additive weight images. A straightforward training process would be directly
using the same process as done for training a vanilla NeRF, only with the above change on the loss function. While such a straightforward training strategy could work, we find that its reconstruction quality is relatively low.


Based on the fact that our \palenerf and the vanilla NeRF share a majority part of network structures, we propose a two-stage \emph{pretrain + finetune} training strategy. Specifically, we first pretrain a vanilla NeRF using RGB input images. After that, we fix the parameters of the density-related layers, i.e. the parameters in the first 9 layers are kept unchanged, and finetune the parameters of the last 2 color-related layers by minimizing L2 differences between rendered and the GT $K$-channel additive weight images. Experiments in Sec.~\ref{sec:ablation} show that such a two-stage training strategy leads to better reconstruction accuracy.

\begin{figure}
	\centering
	\setlength\tabcolsep{1pt}
	\setlength{\imgw}{0.24\linewidth}
	\newcommand{\palew}{2.5mm}
	\begin{tabular}{ccccc}
		& NeRF & KiloNeRF & NeRF & KiloNeRF \\
		\raisebox{0.5\imgw}{\rotatebox[origin=c]{90}{Original}}
		& \includegraphics[width=\imgw]{res/our/lego/004.jpg}
		& \includegraphics[width=\imgw]{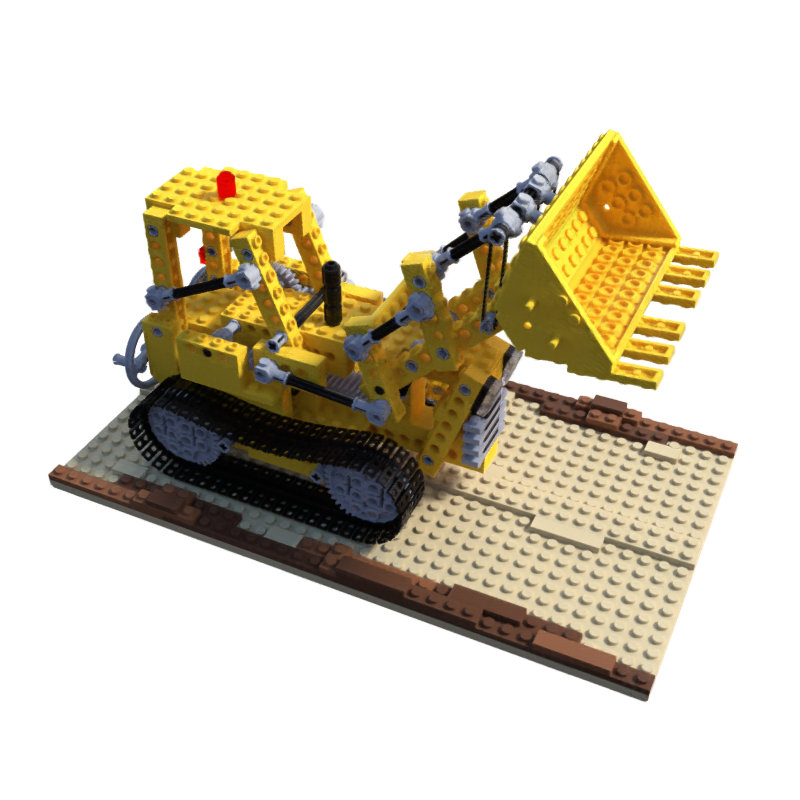}
		& \includegraphics[width=\imgw]{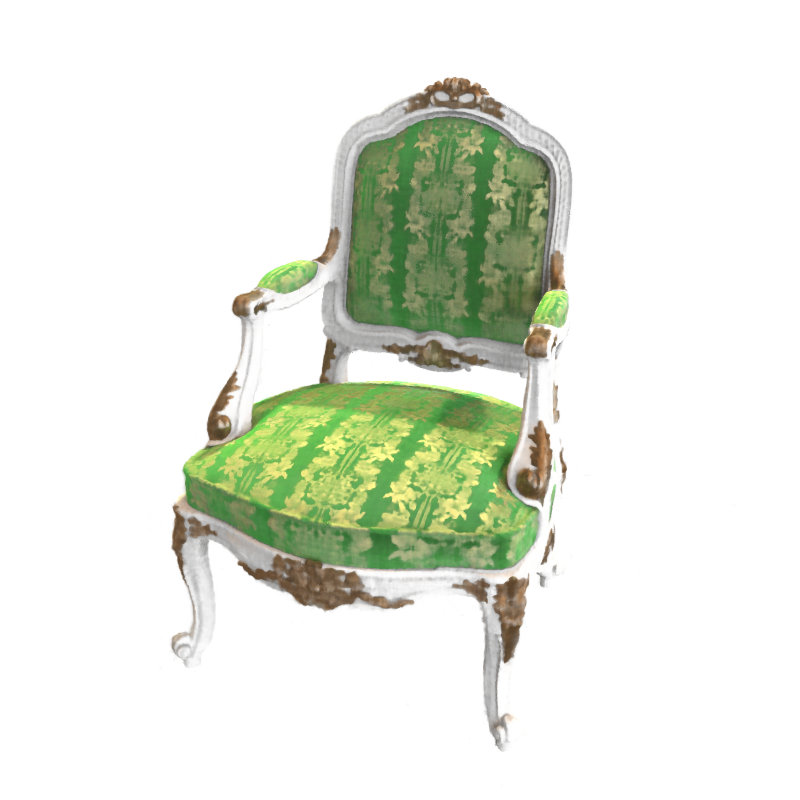}
		& \includegraphics[width=\imgw]{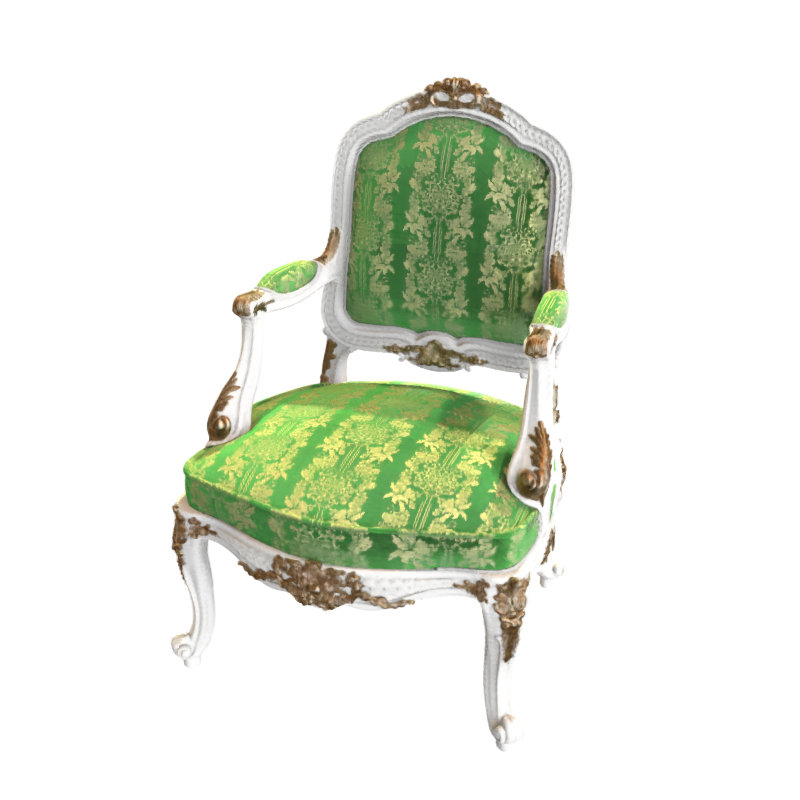}
		\vspace{-1mm} \\
		\raisebox{0.5\imgw}{\rotatebox[origin=c]{90}{Edited}}
		& \includegraphics[width=\imgw]{res/our/lego/004-e0.jpg}
		& \includegraphics[width=\imgw]{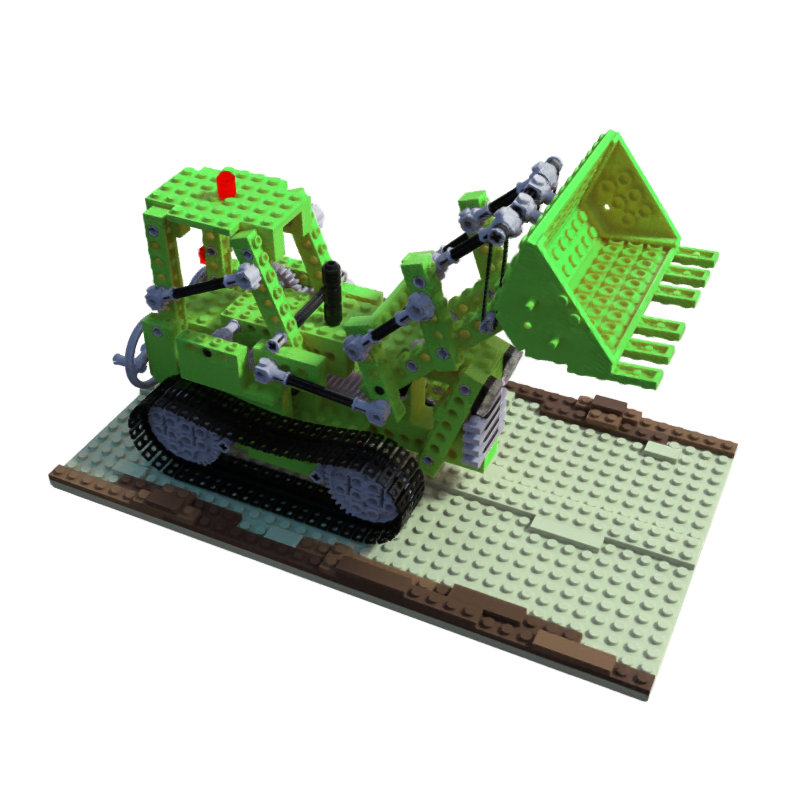}
		& \includegraphics[width=\imgw]{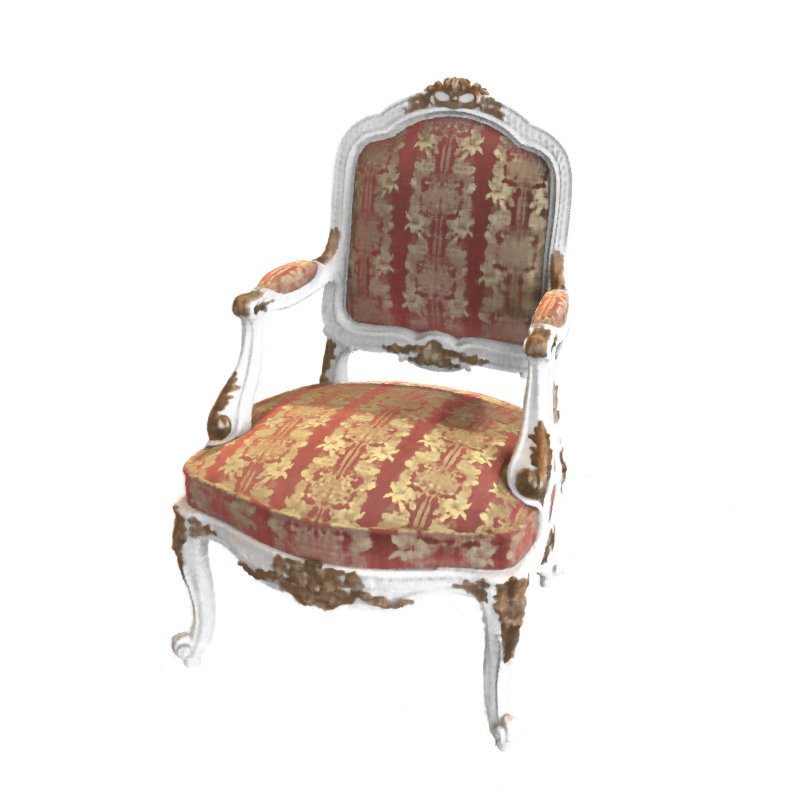}
		& \includegraphics[width=\imgw]{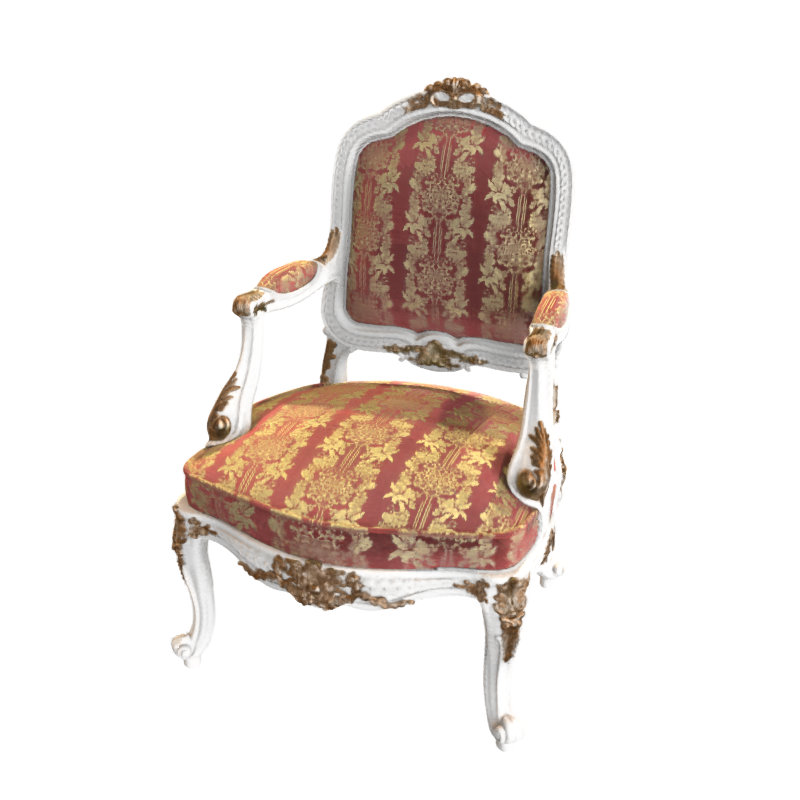}
		\vspace{-1mm} \\
		&
		\multicolumn{2}{c}{
			\begin{tikzpicture}[x=\palew,y=\palew]
				\node[] at (0,2)
				{\adjustbox{frame}{\includegraphics[height=\palew]{res/our/lego/pale.png}}};
				\node[] at (0,0)
				{\adjustbox{frame}{\includegraphics[height=\palew]{res/our/lego/pale-e0.png}}};
				\draw[->, semithick] (2,1.5) -- (2,0.5);
			\end{tikzpicture}
		}
		&
		\multicolumn{2}{c}{
			\begin{tikzpicture}[x=\palew,y=\palew]
				\node[] at (0,2)
				{\adjustbox{frame}{\includegraphics[height=\palew]{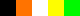}}};
				\node[] at (0,0)
				{\adjustbox{frame}{\includegraphics[height=\palew]{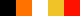}}};
				\draw[->, semithick] (1,1.5) -- (1,0.5);
				\draw[->, semithick] (2,1.5) -- (2,0.5);
			\end{tikzpicture}
		}
	\end{tabular}
	\caption{Editing results of using vanilla NeRF and KiloNeRF as backbones, respectively. Both methods can fit the multi dimensional weights well, and achieve editing results with visually indistinguishable quality.}
	\label{fig:port}
	\end{figure}
\input{fig/res-render}

\section{Method Extensions}


Our method can be easily extended in several ways. In the following, first, we show that the backbone of \palenerf can be replaced by KiloNeRF~\cite{reiser_kilonerf_2021} to achieve real-time recoloring (Sec.~\ref{sec:realtime-recolor}). Second, we extend the palette and mixing weights of \palenerf to capture indirect lighting in synthetic scenes (Sec.~\ref{sec:higher-order weights}). Finally, we would like to show that applications in palette-based editing can be directly incorporated into our framework. Specifically, we demonstrate an application of color harmonization with \palenerf (Sec.~\ref{sec:color harmo}).

\subsection{Real-time Color Editing} \label{sec:realtime-recolor}
As mentioned earlier, our editing process is decoupled with the underlying NeRF backbone. The vanilla NeRF backbone could be replaced by any newer, faster variants of NeRFs as long as they are still volumetric representations. To demonstrate this ability, we also implement our \palenerf with a KiloNeRF backbone \cite{reiser_kilonerf_2021}, which allows color editing in real time. Similar to NeRF, we modify the last FC layer of the small MLP used in KiloNeRF from $32\times 3$ to $32\times K$. This only slows down the rendering time for each frame by around 4\% for $K=5$ and still achieves real-time framerates.
The editing results of the KiloNeRF backbone are shown in Figure \ref{fig:port}. The results with KiloNeRF backbones are visually indistinguishable from those with vanilla NeRF backbones.

\subsection{Second-order weights for synthetic scenes} \label{sec:higher-order weights}

NeRF scenes could be classified into captured scenes and synthetic scenes. Captured scenes refer to those scenes where only captured images from specific views are available. While the images of synthetic scenes are generated using renderers with given 3D geometries, materials, and lighting information. 

For synthetic NeRF scenes, we could further extend our method to provide a more semantically meaningful palette and better controls on recoloring, i.e., supporting higher-order effects (indirect lighting) besides additive mixing. 
Let's imagine a synthetic 3D scene with static geometry and static lighting, but with some editable materials, whose diffuse colors $\vvv_i$ are allowed to adjust. Considering indirect illuminations up to 2 bounces, the rendered color $\vc$ of a pixel $\vp$ could be computed as follows: 
\begin{equation} \label{equ:render_palette_decomposition}
    \vc =  w_0^\vp + \sum_{1\leq i \leq K} w_i^\vp \cdot \vvv_i 
    + \sum_{1\leq i\leq j \leq K} w_{i,j}^\vp \cdot (\vvv_i \odot \vvv_j),
\end{equation}
where $\odot$ denotes channel-wise multiplication, $w_0^\vp$ denotes contributions from light paths not intersecting with any editable materials. $w_i^\vp$ and $w_{i,j}^\vp$ denote contributions from light paths intersecting one and two times of editable materials, respectively. All these weights are essentially light transports and can be directly computed using a path tracing renderer such as PBRT~\cite{pbrt-book-2016}.

The equation can be written in the vectorized form: 
\begin{equation} \label{equ:render_palette_decomposition2}
    \vc = \MV \cdot \vw^\vp,
\end{equation}
where:
\begin{equation} \label{equ:render_weights}
    \begin{array}{cl} 
        \MV = & [\mathbf{1},\vvv_1,\dots,\vvv_K,\vvv_{1,1},\dots,\vvv_{K,K}],   \\ 
        \vw^\vp =  & [w_0^\vp,w_1^\vp, \dots, w_K^\vp,w_{1,1}^\vp,\dots,w_{K,K}^\vp]^T.
    \end{array}
\end{equation}
The above modification could be directly supported by our \palenerf, by only modifying the data preparation step at the beginning and the color editing step at the end, as described in Sec.~\ref{sec:overview}. Notice that a palette size of $K$ would produce a second-order weight vector $\MV$ of length $(K+1)(K+2)/2$.

In the data preparation step,  we no longer use existing palette-based image decomposition methods to extract palette and additive weights. Instead, we ask users to manually specify several materials he or she wants to adjust and use the diffuse colors of specified materials as palette colors. Then, we compute the second-order weights in Eq.~\ref{equ:render_weights} directly using PBRT~\cite{pbrt-book-2016}. In the color editing step, we use Eq.~\ref{equ:render_palette_decomposition} instead of Eq.~\ref{equ:palette_decomposition} to compute the reconstructed colors.


We have tested two synthetic scenes: \emph{Cornell box} and \emph{breakfast}. The editing results are shown in Figure~\ref{fig:render-img}. Notice the visually plausible indirect illumination effects after recoloring, e.g., color bleeding effects in the ceiling and reflection of the sphere on the cylinder surface in scene \emph{Cornell box}, and color bleeding effects including pink tint of the floor and green tint of the desk in scene \emph{breakfast}. 

\begin{figure}[t]
    \small
	\setlength\tabcolsep{1pt}
	\renewcommand{\arraystretch}{0.7}
    \setlength{\imgw}{0.33\linewidth}
    \newcommand{\palew}{3mm}
    \newcommand{\wheelw}{21mm}
    \centering
    \begin{tabular}{ccc}
        Original & Monochrome & Complementray \\
        \includegraphics[width=\imgw]{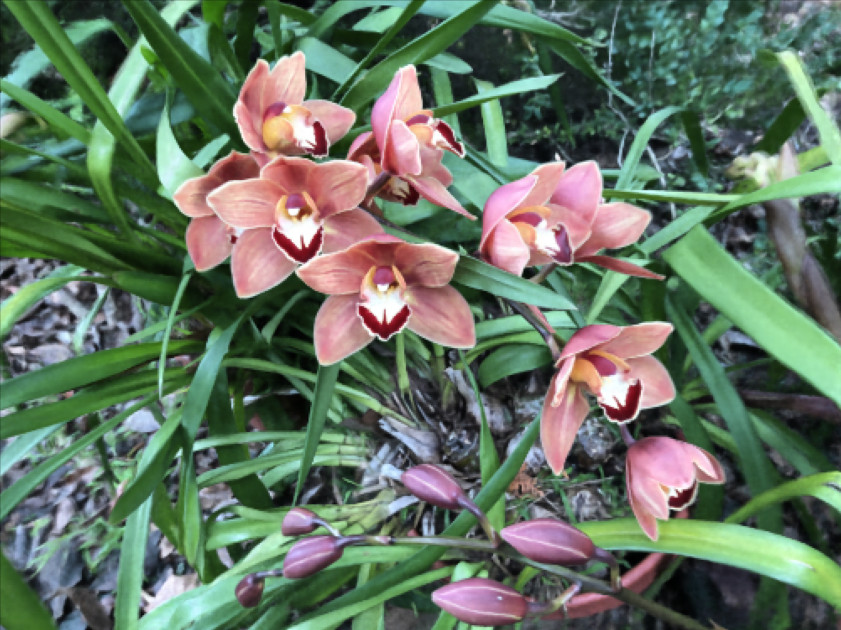} &
        \includegraphics[width=\imgw]{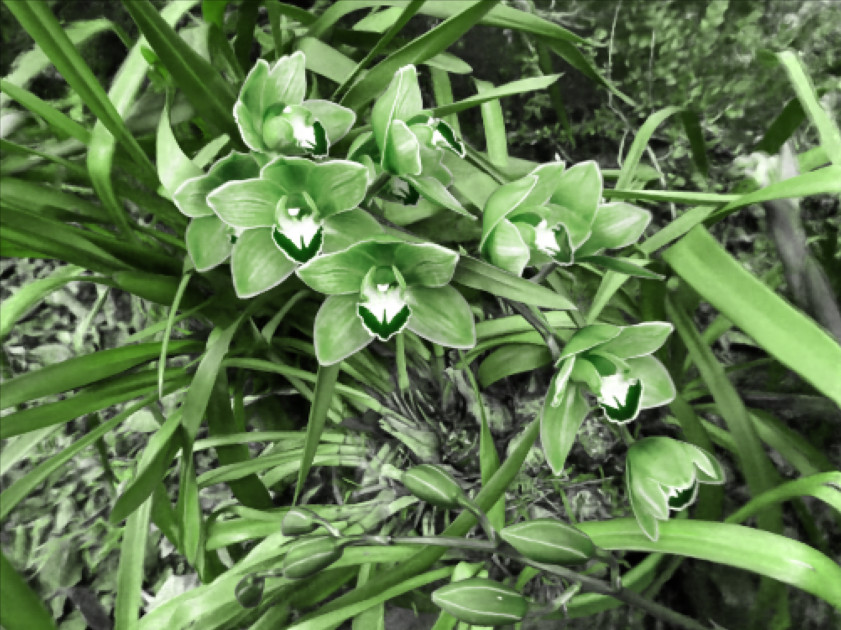} &
        \includegraphics[width=\imgw]{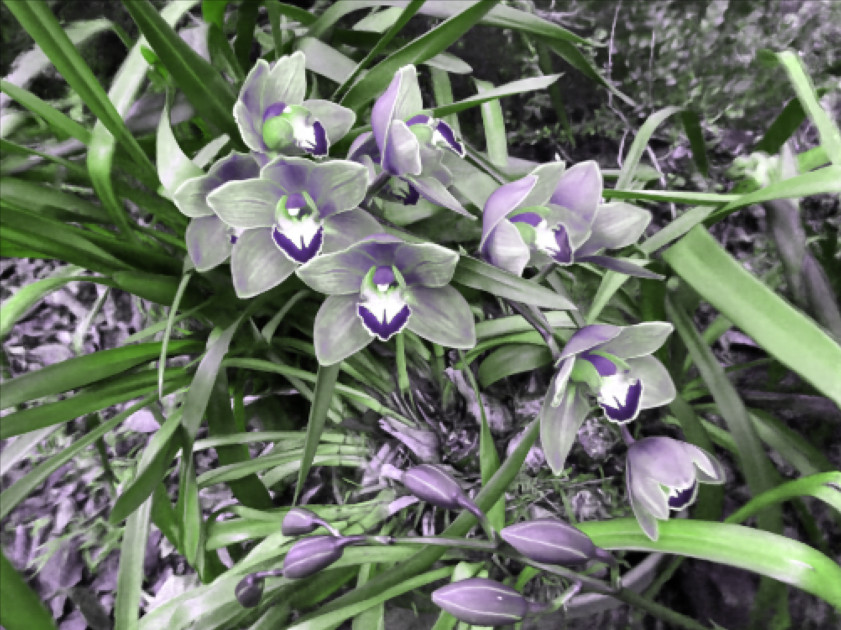}
        \\
        \adjustbox{frame}{\includegraphics[height=\palew]{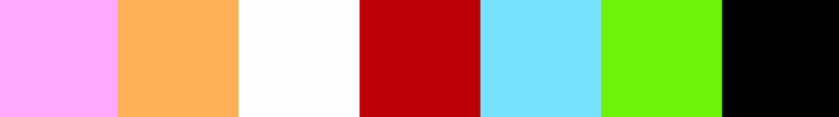}} &
        \adjustbox{frame}{\includegraphics[height=\palew]{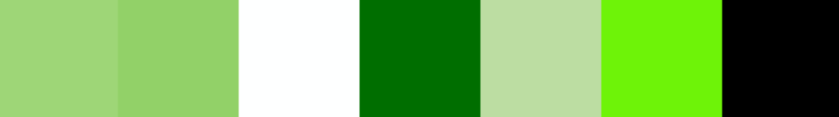}} &
        \adjustbox{frame}{\includegraphics[height=\palew]{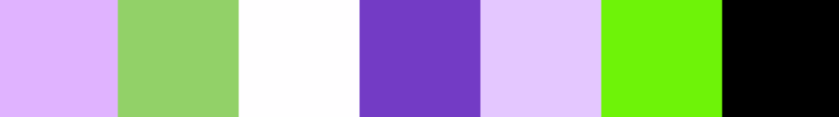}}
        \\
        \adjustbox{frame}{\includegraphics[width=\wheelw]{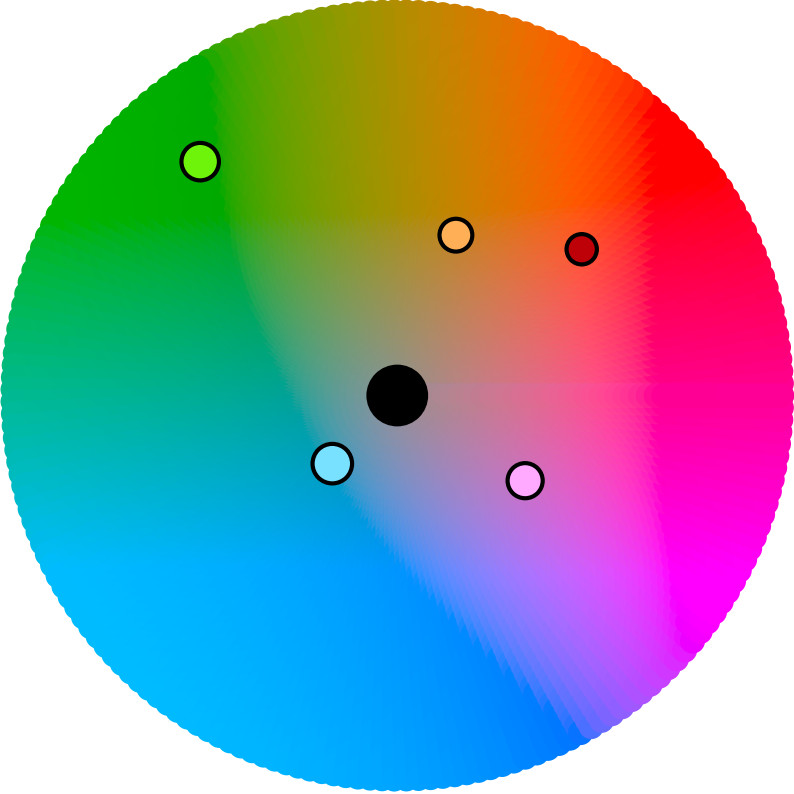}} &
        \adjustbox{frame}{\includegraphics[width=\wheelw]{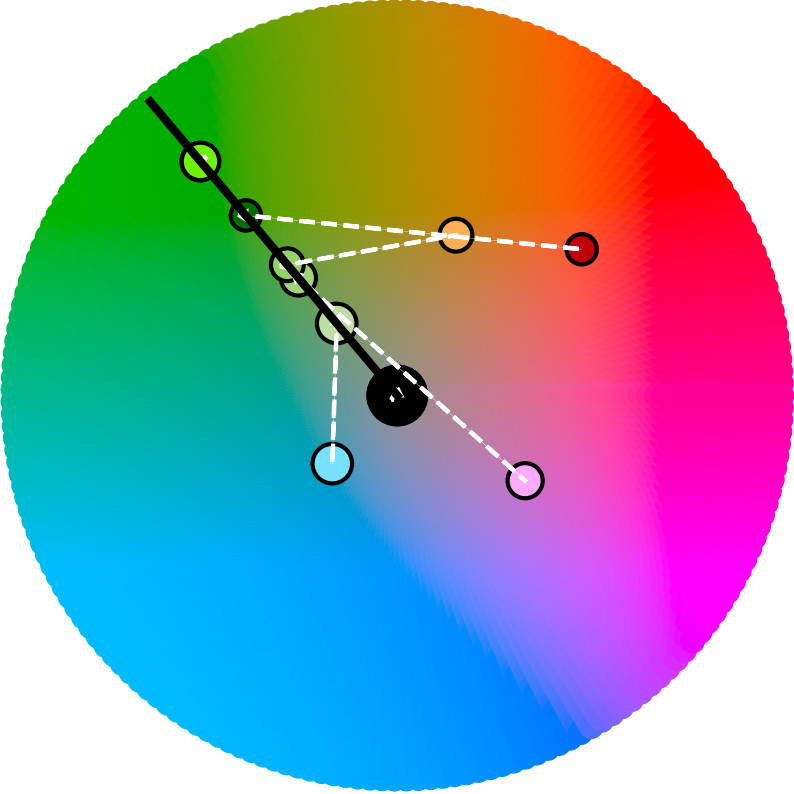}} &
        \adjustbox{frame}{\includegraphics[width=\wheelw]{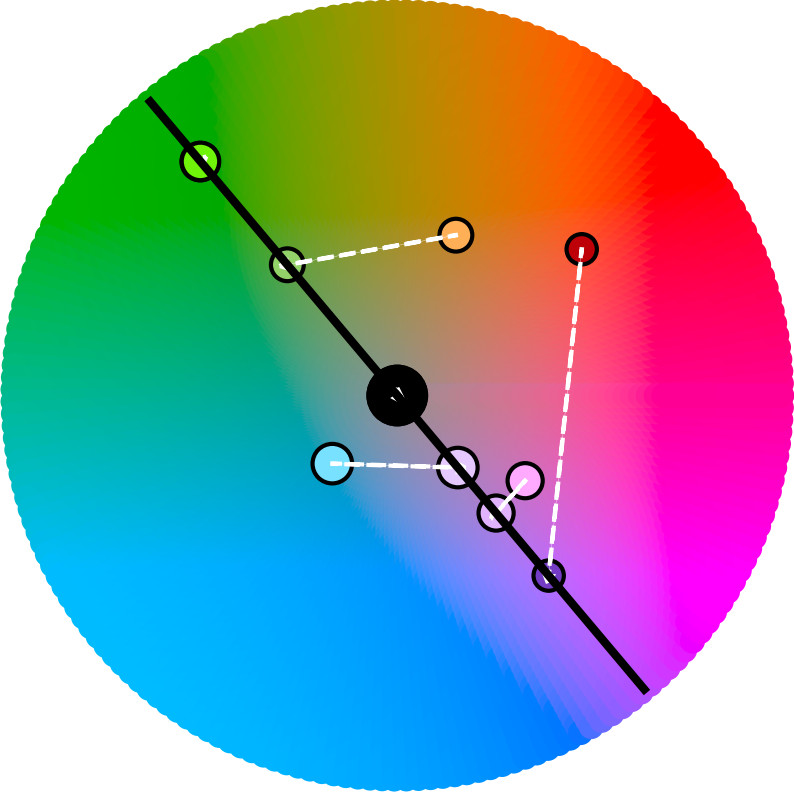}}

        \\
        \specialrule{0em}{0pt}{5pt}
        Triad & Square & Analogous \\
        \includegraphics[width=\imgw]{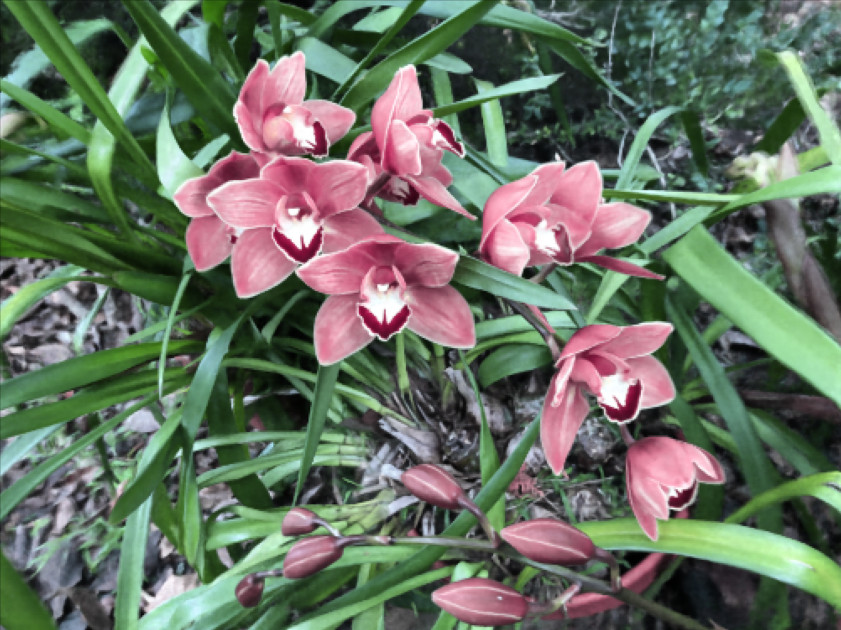} &
        \includegraphics[width=\imgw]{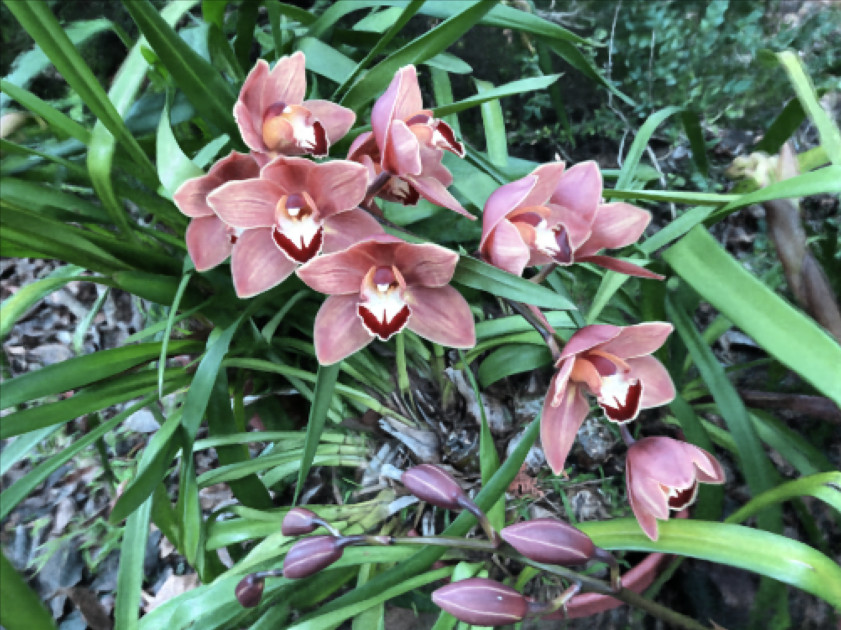} &
        \includegraphics[width=\imgw]{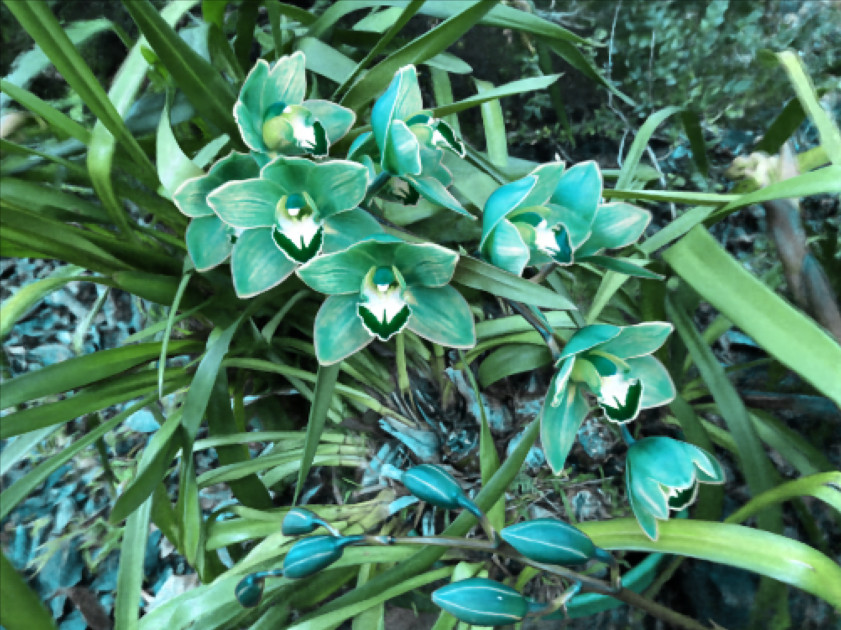}
        \\
        \adjustbox{frame}{\includegraphics[height=\palew]{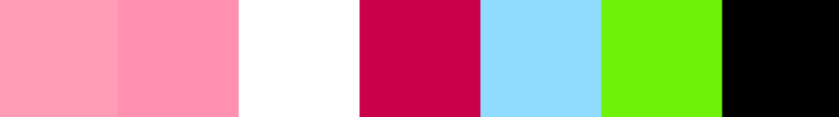}} &
        \adjustbox{frame}{\includegraphics[height=\palew]{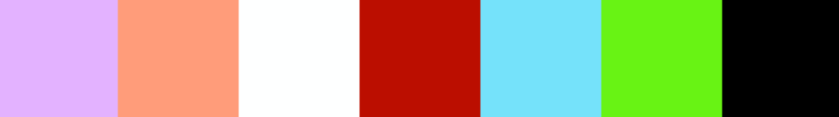}} &
        \adjustbox{frame}{\includegraphics[height=\palew]{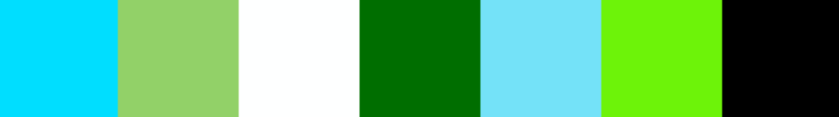}}
        \\
        \adjustbox{frame}{\includegraphics[width=\wheelw]{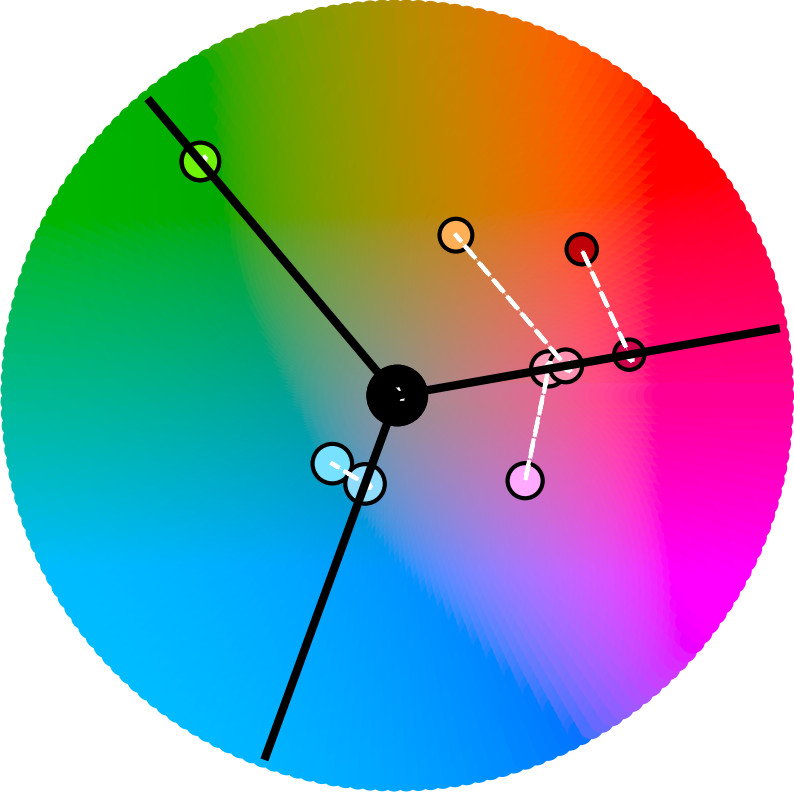}} &
        \adjustbox{frame}{\includegraphics[width=\wheelw]{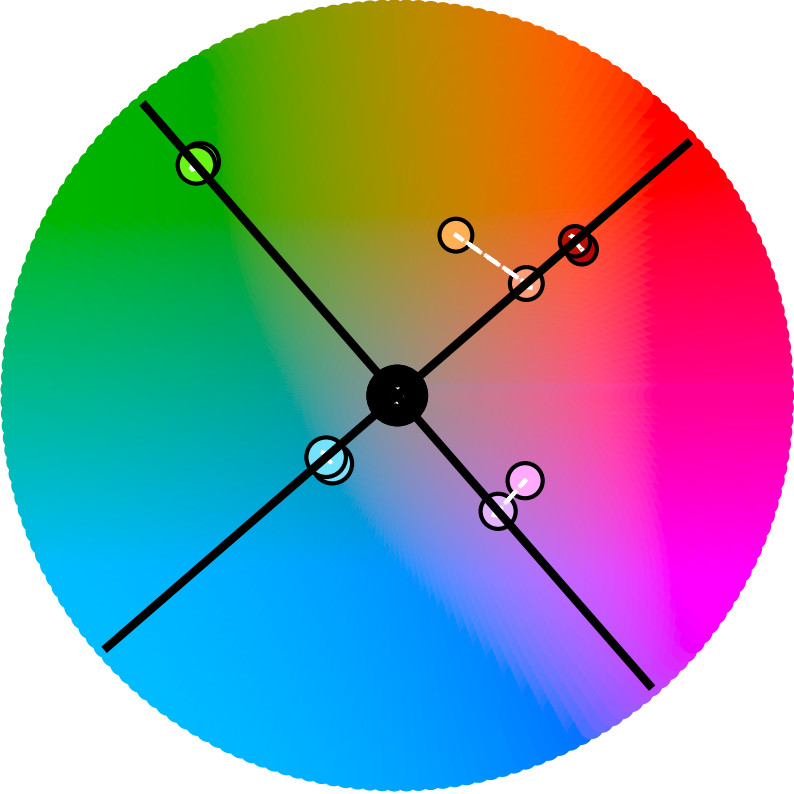}} &
        \adjustbox{frame}{\includegraphics[width=\wheelw]{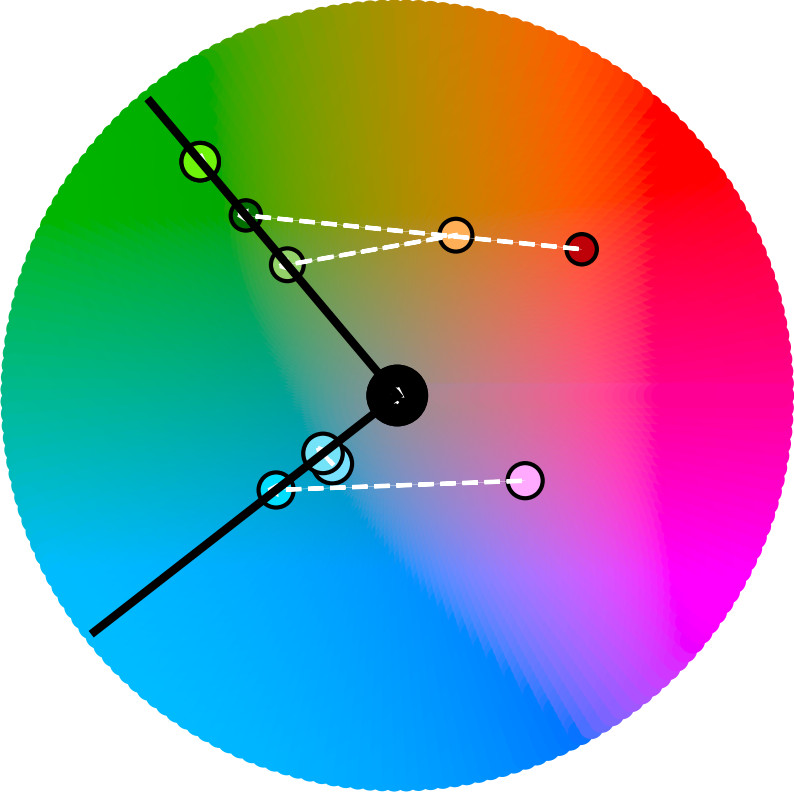}}

        \\
        \specialrule{0em}{0pt}{5pt}
        Single Split & \makecell[c]{Double Split\\(Optimal)} & \\
        \includegraphics[width=\imgw]{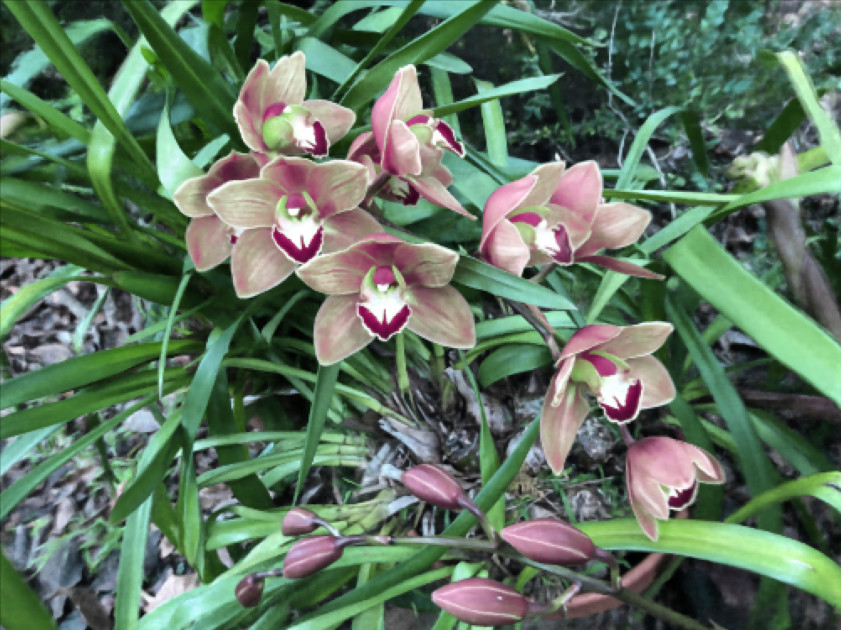} &
        \includegraphics[width=\imgw]{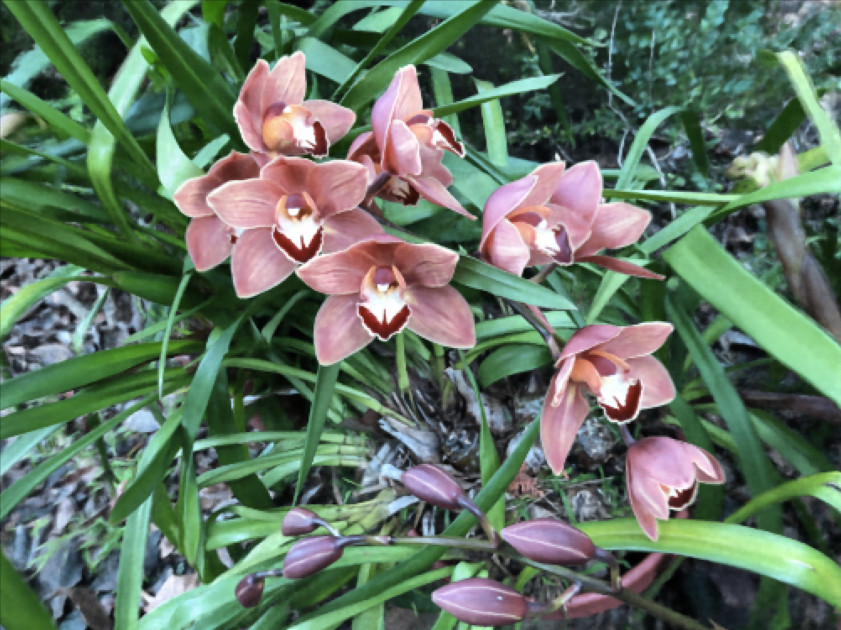} &
        \\
        \adjustbox{frame}{\includegraphics[height=\palew]{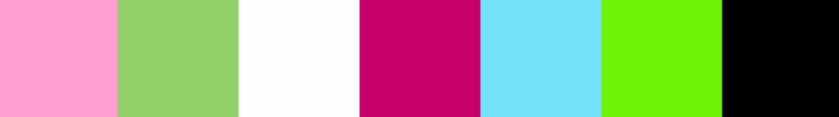}} &
        \adjustbox{frame}{\includegraphics[height=\palew]{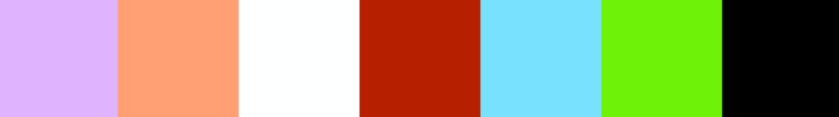}} &
        \\
        \adjustbox{frame}{\includegraphics[width=\wheelw]{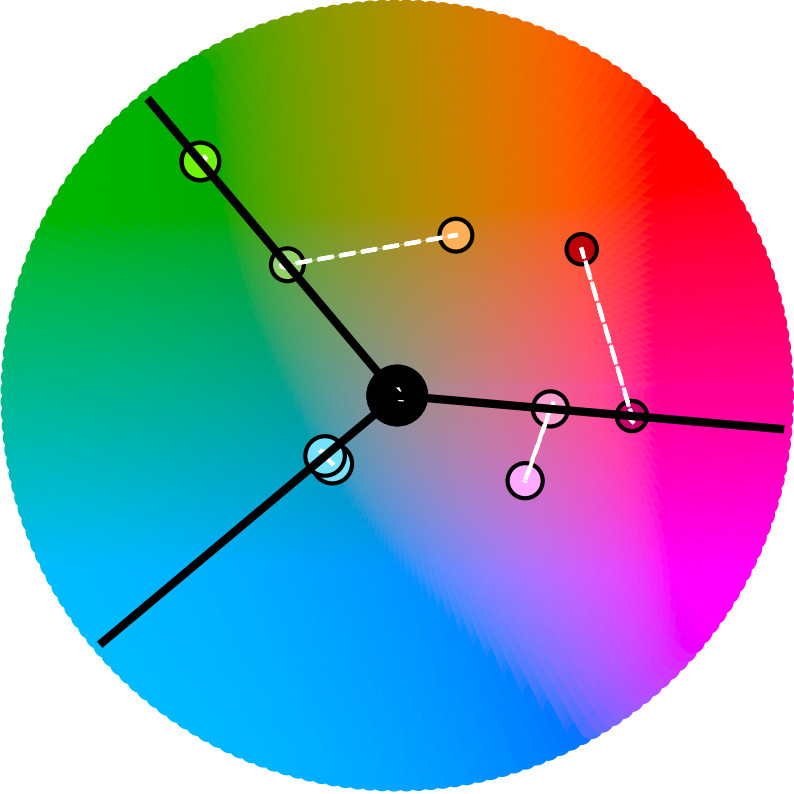}} &
        \adjustbox{frame}{\includegraphics[width=\wheelw]{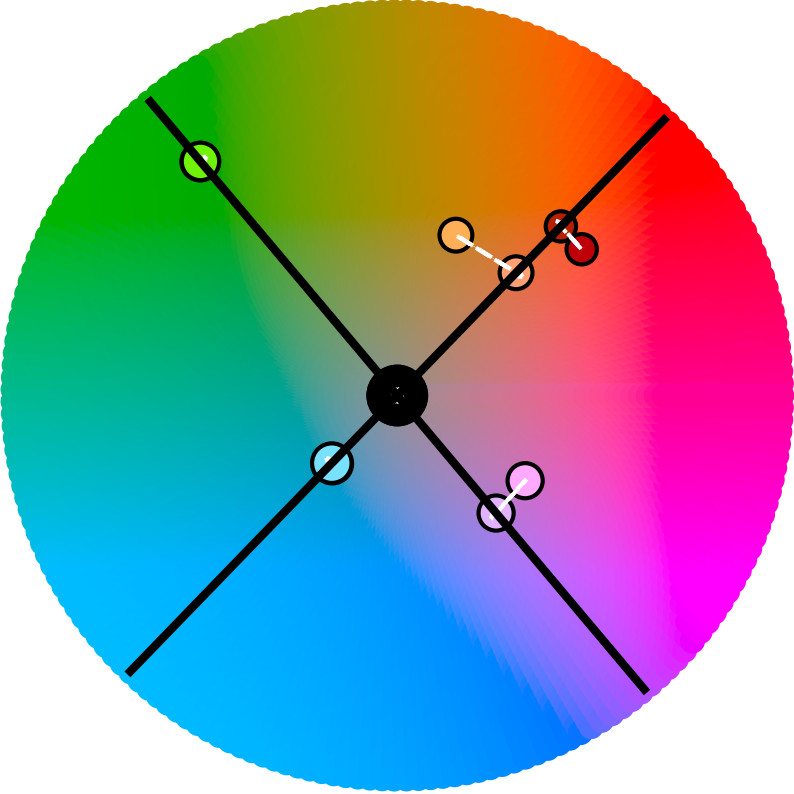}}
    \end{tabular}
    
    \caption{Comparison of the same scene \emph{orchids} harmonized by 7 different templates.
    }
    \label{fig:diff-templates}
\end{figure}

\subsection{Color Harmonization} \label{sec:color harmo}

Our method could be directly combined with the palette-based image harmonization method~\cite{tan_palette-based_2018} to achieve color harmonization for NeRFs. 

Specifically, we use the 7 harmonic color templates $T_m, (m=1\ldots 7$) designed for palette colors in \cite{tan_palette-based_2018}. Each harmonic template has at least 1 parameter $\alpha$, which describes the angle of rotation in the hue of the axis. 
The harmonization is performed in HCL color space (Hue, Chroma, Luminance). 
We utilize the palette $\MV$ extracted in the data preparation step of \palenerf. For template $T_m(\alpha)$, we find the closest axis to each color in $\MV$, measured by the distance of hue in HCL color space. The distance is weighted and summed to compute a template fitting cost, which is then minimized to find the best $\alpha$ for each template. 

Once the template is chosen, each palette color in $\MV$ is projected onto the nearest axis of the template, forming the harmonized palette $\MV'$. Then, as described in Sec.~\ref{sec:overview}, $\MV'$ is used to recolor, i.e. harmonize the whole scene.
Figure~\ref{fig:diff-templates} shows harmonized results of scene \emph{orchids}.

\section{Experiments}

We conduct our experiments on a PC with an NVIDIA RTX 3080Ti GPU, a Ryzen 5900X CPU, and 64GB of RAM. For each scene, we train \palenerf for 200k iterations, which take around 5 hours.

\subsection{Evaluation} \label{sec:ablation}

\begin{table}[tbp]
\small
\setlength\tabcolsep{2.5pt}
\begin{tabular}{cllllll}
\toprule
 \multirow{2}{*}{\begin{tabular}[c]{@{}c@{}} training\\method\end{tabular}} 
 & \multicolumn{3}{c}{Additive Weights} & \multicolumn{3}{c}{Second Order Weights} \\
 & PSNR↑       & SSIM↑     & LPIPS↓     & PSNR↑        & SSIM↑       & LPIPS↓      \\
\cmidrule(lr){1-1} \cmidrule(lr){2-4} \cmidrule(lr){5-7}
traditional & 25.205 & 0.832 & 0.129 & 8.579  & 0.084 & 0.363 \\
ours        & 26.711 & 0.833 & 0.095 & 40.107 & 0.976 & 0.024 \\
\bottomrule
\end{tabular}

\vspace{1mm}
\caption{Quantitative comparison between the traditional strategy and our two-stage (pretrain + finetune) training strategy. ``Additive Weights'' columns report average scores of 6 captured scenes (\emph{lego, chair, flower, orchids, vasedeck, valley}). ``Second Order Weights'' columns report average scores of 2 synthetic scenes (\emph{Cornell box, breakfast}). }
\label{tab:ablation-pretrain}

\end{table}
\begin{table}[tbp]
\small
\setlength\tabcolsep{3pt}
\centering
\begin{tabular}{lllllll}
\toprule
            & \multicolumn{3}{c}{vasedeck} & \multicolumn{3}{c}{lego} \\ 
dimension   & PSNR↑    & SSIM↑   & LPIPS↓  & PSNR↑   & SSIM↑ & LPIPS↓ \\
\cmidrule(lr){1-1} \cmidrule(lr){2-4} \cmidrule(lr){5-7}
128         & 24.262   & 0.693   & 0.172   & 30.297  & 0.941 & 0.031  \\
256         & 24.730   & 0.722   & 0.154   & 30.525  & 0.944 & 0.030  \\
512         & 24.732   & 0.720   & 0.156   & 30.664  & 0.945 & 0.029  \\
NeRF(128)   & 24.713   & 0.715   & 0.163   & 30.604  & 0.946 & 0.030  \\
\bottomrule
\end{tabular}

\vspace{1mm}
\caption{Ablation studies on the dimension of the last layer of \palenerf (shown in the first 3 rows) evaluated on scenes \emph{vasedeck} and \emph{lego}. The last row shows scores of the vanilla NeRF with 128 dimension for comparison.
}
\label{tab:ablation-dim}
\end{table}

\paragraph{Training strategy.}
As mentioned in Sec.~\ref{sec:network_and_training}, we have proposed a two-stage (pretrain + finetune) training strategy. To validate the effectiveness of the two-stage strategy, we compare it against the traditional one.  For a fair comparison, we set the number of iterations to be the same for both training strategies. Specifically, the traditional strategy uses 200k iterations, while our two-stage strategy set iteration steps as 180k (pretrain) + 20k (finetune) for captured scenes, and 140k (pretrain) + 60k (finetune) for synthetic scenes. 
We have tested 8 scenes, including 6 captured scenes and 2 synthetic scenes, and report the average reconstruction scores for both strategies in Table \ref{tab:ablation-pretrain}.

From the results, we could find that our two-stage training strategy improves the reconstruction quality by a large gap, especially on the synthetic scenes with second-order weights. This is possibly due to the relatively large number of channels making it harder to be directly trained, while a pretrain step helps provide a good initialization. 



\paragraph{Dimension of the last layer.} We also evaluate different choices of the input dimension of the last layer. 
Table \ref{tab:ablation-dim} shows the reconstruction scores on two scenes when the dimension of the last layer is set to 128, 256, and 512, respectively. Increasing the dimension of the last layer could slightly increase reconstruction quality, at the cost of longer rendering time. 
However, to make a good trade-off between quality and speed, we set it as 128 in our experiments.

\subsection{Comparisons}

\input{fig/compare-poster}
\begin{figure}[t]
\centering
\setlength\tabcolsep{1pt}
\setlength{\imgw}{0.33\linewidth}
\newcommand{\palew}{2.3mm}
\begin{tabular}{ccc}
    Source & CLIP-NeRF & Ours \\
    \includegraphics[width=\imgw]{res/our/lego/004.jpg} &
    \includegraphics[width=\imgw]{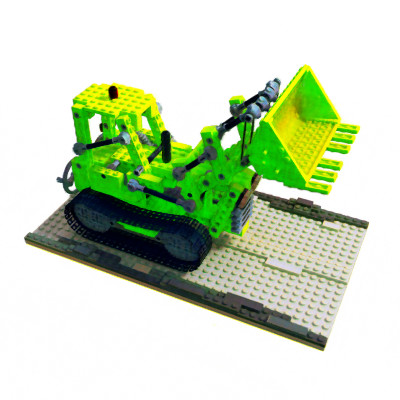} &
    \includegraphics[width=\imgw]{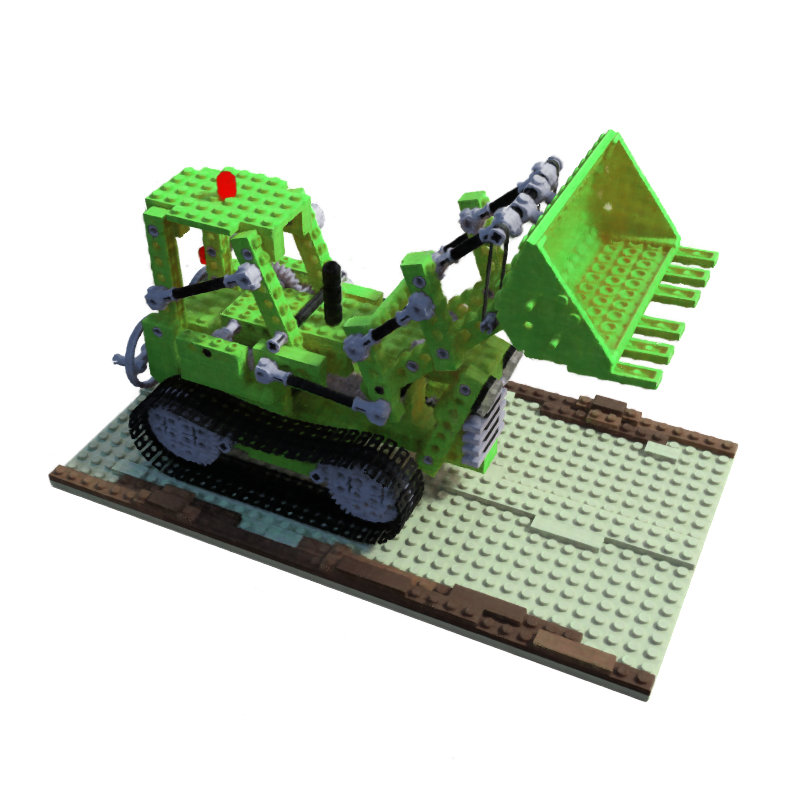} 
    \vspace{-3mm} \\
    &
    \raisebox{6mm}{``A green excavator''} &
    \begin{tikzpicture}[x=\palew,y=\palew]
        \node[] (before) at (0,2)
        {\adjustbox{frame}{\includegraphics[height=\palew]{res/our/lego/pale.png}}};
        \node[] (after) at (0,0)
        {\adjustbox{frame}{\includegraphics[height=\palew]{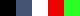}}};
        \draw[->, semithick] (2,1.5) -- (2,0.5);
    \end{tikzpicture}
\end{tabular}

\caption{Comparison with CLIP-NeRF~\cite{wang_clip-nerf_2022}, which takes ``A green excavator'' as text edit prompt to edit the scene. We change one of the palette colors from yellow to green.}
\label{fig:compare-clip}
\end{figure}
\begin{table*}[tbp]
\setlength\tabcolsep{4pt}
\centering
\begin{tabular}{lllllllllllll}
\toprule
            & \multicolumn{3}{c}{lego}     & \multicolumn{3}{c}{chair}  & \multicolumn{3}{c}{flower}      & \multicolumn{3}{c}{orchids}   \\
method    & PSNR↑     & SSIM↑  & LPIPS↓  & PSNR↑    & SSIM↑  & LPIPS↓ & PSNR↑     & SSIM↑    & LPIPS↓   & PSNR↑     & SSIM↑   & LPIPS↓  \\
\cmidrule(lr){1-1} \cmidrule(lr){2-4} \cmidrule(lr){5-7} \cmidrule(lr){8-10} \cmidrule(lr){11-13} 
NeRF      & 31.153 & 0.952 & 0.025 & 31.322 & 0.956 & 0.041 & 28.147 & 0.887 & 0.053 & 21.325 & 0.750 & 0.109 \\
\palenerf & 30.135 & 0.941 & 0.030 & 30.723 & 0.950 & 0.045 & 28.172 & 0.881 & 0.059 & 21.515 & 0.759 & 0.108 \\
\midrule
            & \multicolumn{3}{c}{vasedeck} & \multicolumn{3}{c}{valley} & \multicolumn{3}{c}{Cornell box} & \multicolumn{3}{c}{breakfast} \\
method    & PSNR↑     & SSIM↑  & LPIPS↓  & PSNR↑    & SSIM↑  & LPIPS↓ & PSNR↑     & SSIM↑    & LPIPS↓   & PSNR↑     & SSIM↑   & LPIPS↓  \\
\cmidrule(lr){1-1} \cmidrule(lr){2-4} \cmidrule(lr){5-7} \cmidrule(lr){8-10} \cmidrule(lr){11-13} 
NeRF      & 24.713 & 0.715 & 0.163 & 27.099 & 0.760 & 0.137 & 44.151 & 0.993 & 0.007 & 39.952 & 0.971 & 0.03  \\
\palenerf & 24.657 & 0.721 & 0.152 & 26.638 & 0.745 & 0.159 & 41.339 & 0.985 & 0.013 & 38.875 & 0.967 & 0.035 \\
\bottomrule
\end{tabular}
\vspace{1mm}
\caption{Quantitative comparison of \palenerf to the vanilla NeRF on 8 scenes. We report PSNR/SSIM (higher is better) and LPIPS (lower is better). \palenerf is able to achieve similar performance compared to the vanilla NeRF.
}
\label{tab:real-metric}
\end{table*}

\paragraph{Comparison with PosterNeRF.} 

Figure~\ref{fig:compare-poster} compares our edited results with a concurrent work -- PosterNeRF~\cite{tojo_poster-nerf_2022} on two scenes \emph{lego} and \emph{drums}. 
PosterNeRF also employs a palette-based editing interface. However, it could easily generate visible artifacts, i.e., producing inconsistent recoloring results. Please see the close-up images in Figure~\ref{fig:compare-poster}: PosterNeRF produces unnatural recoloring on the ground near the excavator in scene \emph{lego} and generates unsmooth recoloring on the cymbal in scene \emph{drums}. In contrast, our recoloring results are much better without artifacts.

The reason why PosterNeRF generates artifacts lies in the way how it uses palettes and computes mixing weights. PosterNeRF will choose at most 2 palette colors, resulting in at most 2 non-zero values of weight $\vw^\vp$ in Eq.~\ref{equ:palette_decomposition2}. Thus some colors can not be reproduced precisely. Furthermore, the weights are also quantized with large step sizes, which leads to color banding artifacts. 
In contrast, our method approximates each pixel with a linear blending of an arbitrary number of palettes, and the weight of linear blending is accurate, thus producing smooth results.

\paragraph{Comparison with CLIP-NeRF.}
In Figure~\ref{fig:compare-clip}, we further compare our color editing results with CLIP-NeRF~\cite{wang_clip-nerf_2022} on scene \emph{lego}. We can see that CLIP-NeRF tends to blur the detail of the excavator, add noise to the ground, and modify undesired regions such as red lights and grey-blue shafts. In contrast, our method has better image quality and provides better controllability for color editing. 

We do not compare with EditingNeRF~\cite{liu_editnerf_2021} since it is hard to make a fair comparison. Our method, like vanilla NeRF, could be trained through a single scene and used to recolor the scene. In contrast, EditingNeRF requires a scene dataset with many instances from the same category to enable editing, disabling its ability to edit a single NeRF scene (such as the standard ones like \emph{lego} and \emph{drums}).



\begin{figure}
\centering
\renewcommand{\imgw}{0.31\linewidth}
\setlength\tabcolsep{1pt}
\begin{tabular}{cccc}
    & NeRF & \palenerf & GT \\
    \rotatebox[origin=c]{90}{\emph{lego}}
    & \raisebox{-0.5\height}{\includegraphics[width=\imgw]{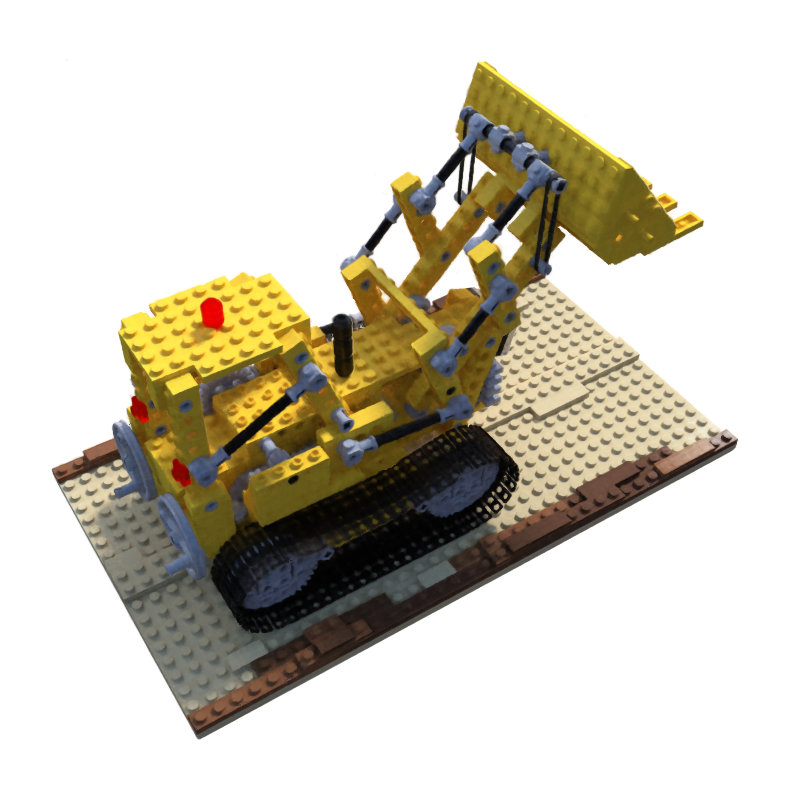}}
    & \raisebox{-0.5\height}{\includegraphics[width=\imgw]{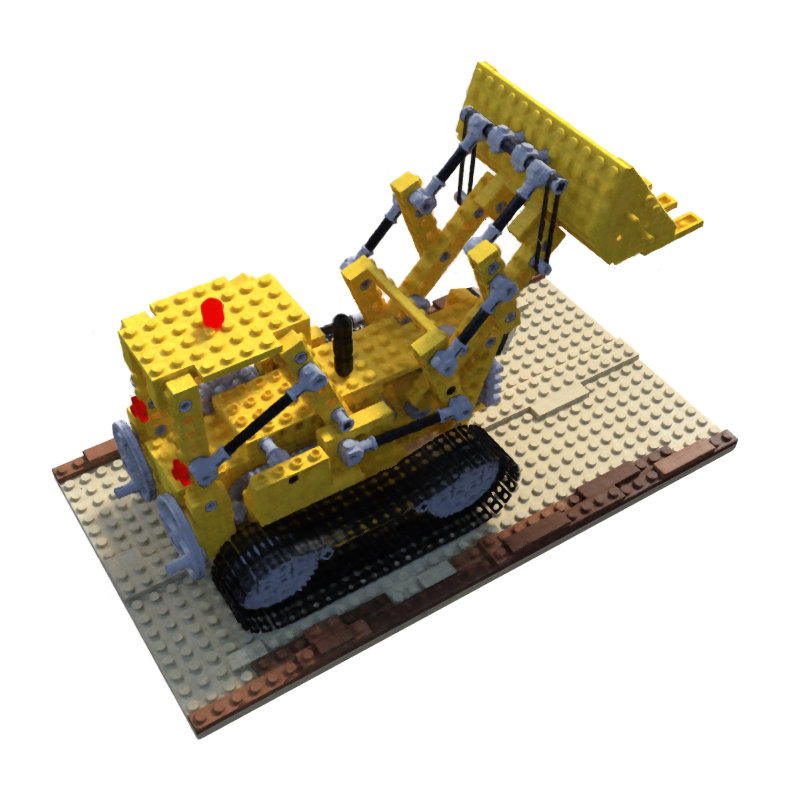}}
    & \raisebox{-0.5\height}{\includegraphics[width=\imgw]{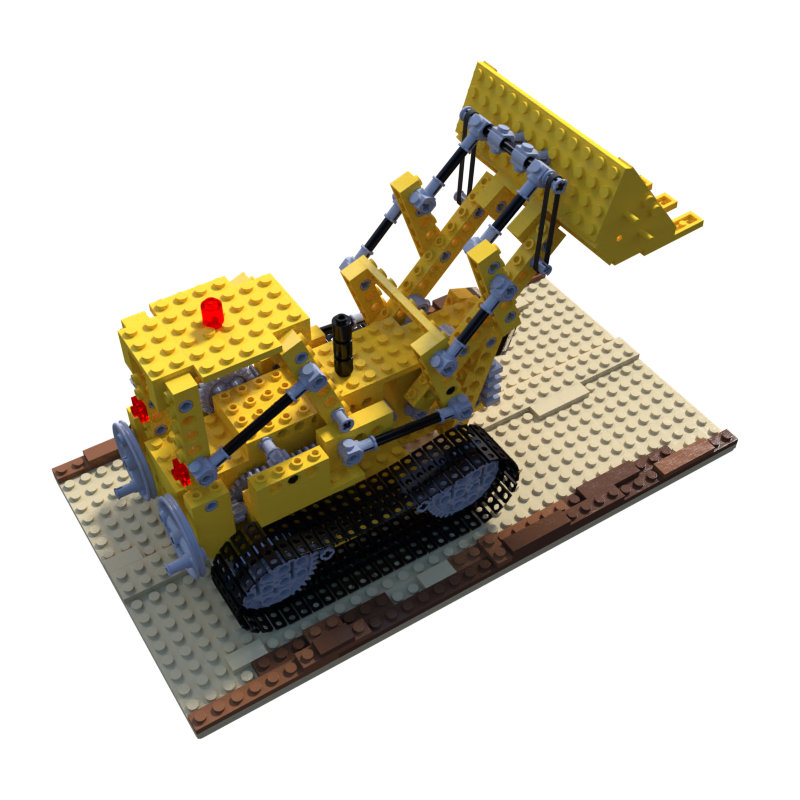}}
    \\
    \rotatebox[origin=c]{90}{\emph{orchids}}
    & \raisebox{-0.5\height}{\includegraphics[width=\imgw]{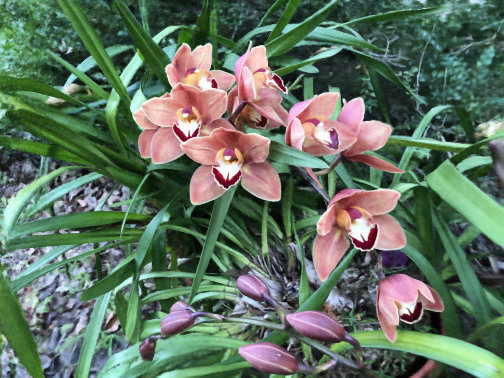}}
    & \raisebox{-0.5\height}{\includegraphics[width=\imgw]{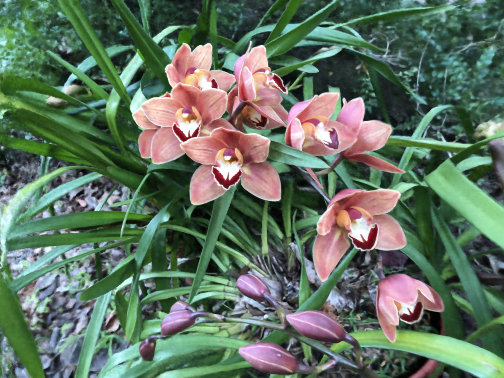}}
    & \raisebox{-0.5\height}{\includegraphics[width=\imgw]{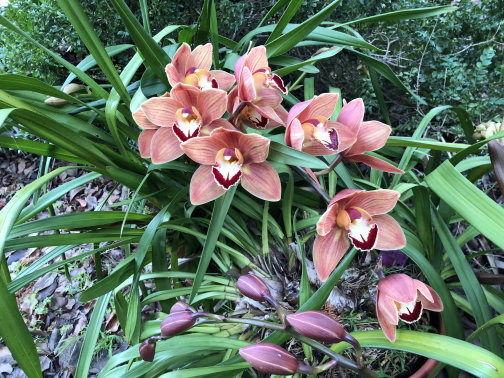}}
\end{tabular}

\caption{Qualitative comparison of \palenerf to the vanilla NeRF. For each scene, from left to right, we show novel views synthesised by the vanilla NeRF and by our method, respectively, and the ground truth.}
\label{fig:compare-real}
\end{figure}

\subsection{Results} 

\begin{figure*}
	\setlength{\imgw}{0.14\linewidth}
	\newcommand{\palew}{2.3mm}
	\setlength\tabcolsep{1pt}
	\renewcommand{\arraystretch}{0.5}
	
	\begin{tabular}{cllllllc}
		\raisebox{0.5\imgw}{chair} & 
		\adjustbox{}{\includegraphics[width=\imgw]{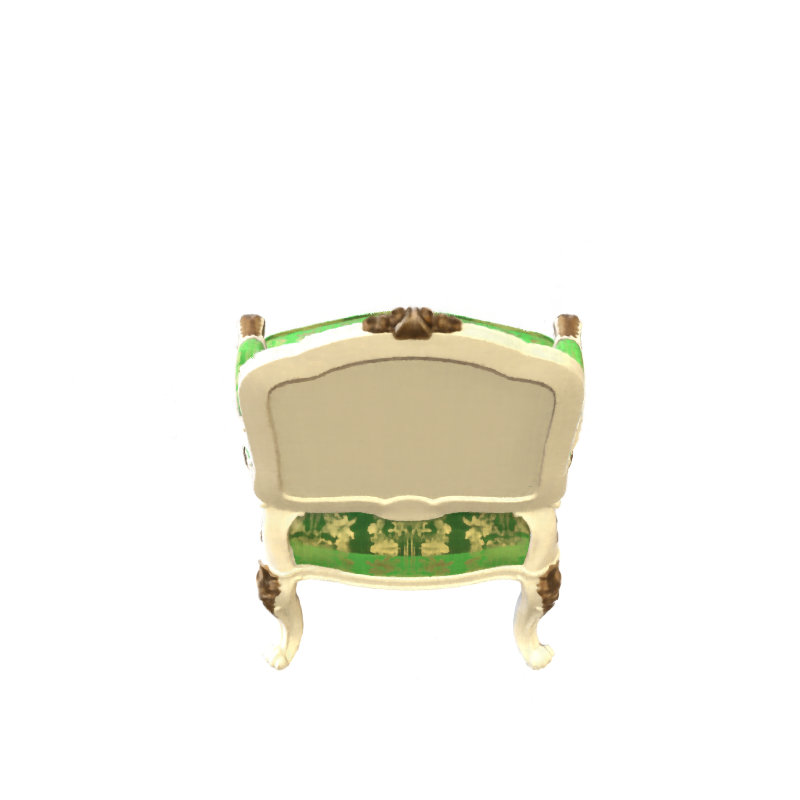}} &
		\adjustbox{}{\includegraphics[width=\imgw]{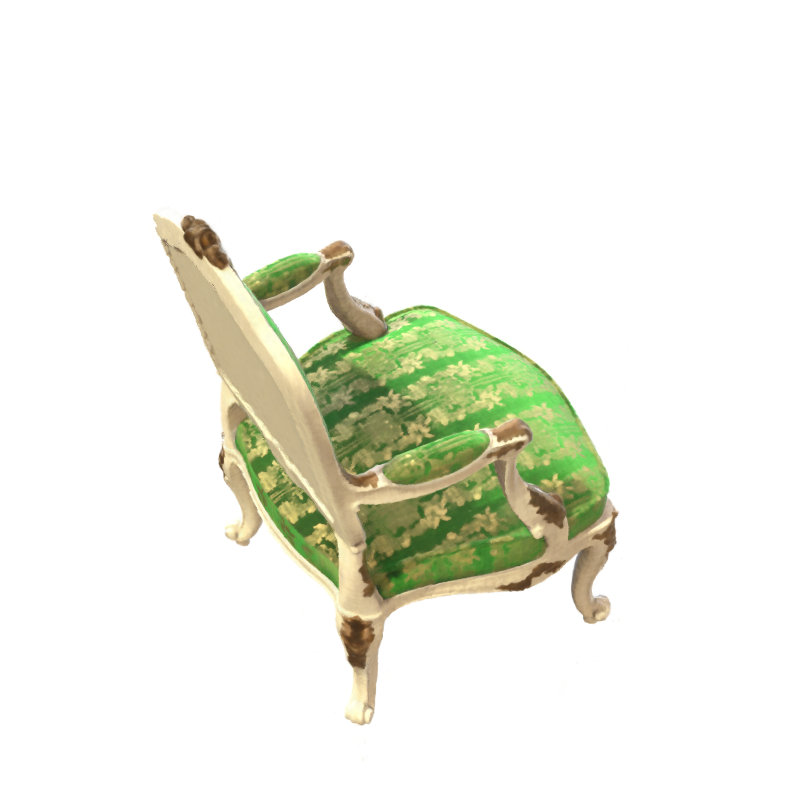}} &
		\adjustbox{}{\includegraphics[width=\imgw]{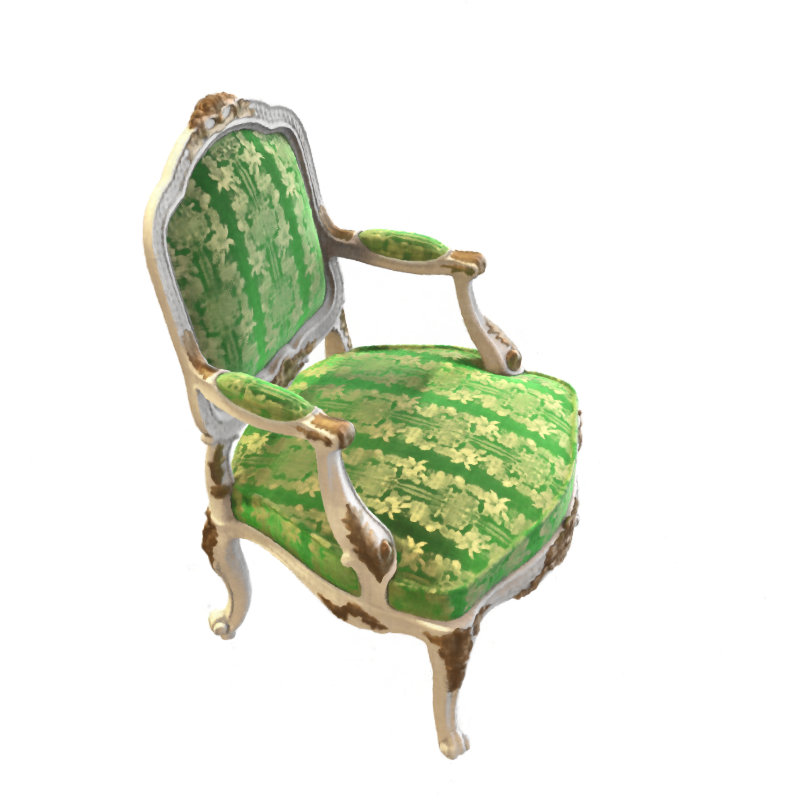}} &
		\adjustbox{}{\includegraphics[width=\imgw]{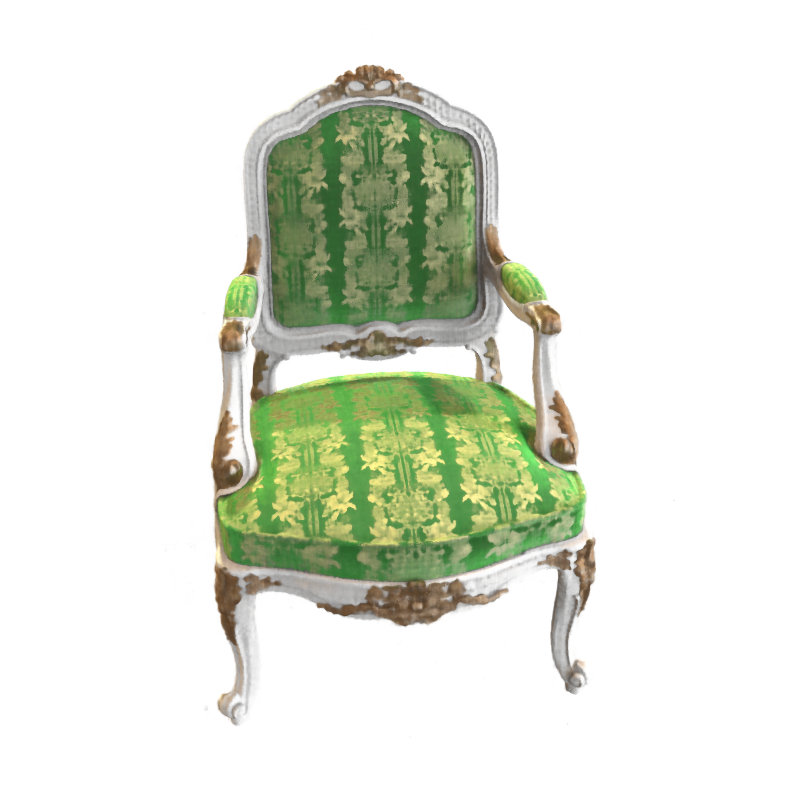}} &
		\adjustbox{}{\includegraphics[width=\imgw]{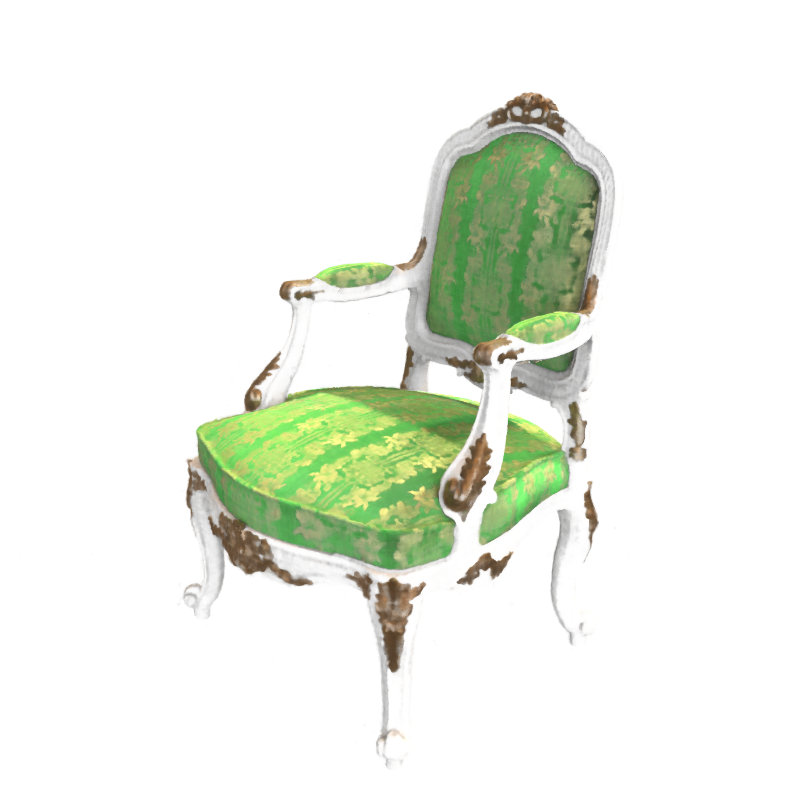}} &
		\adjustbox{}{\includegraphics[width=\imgw]{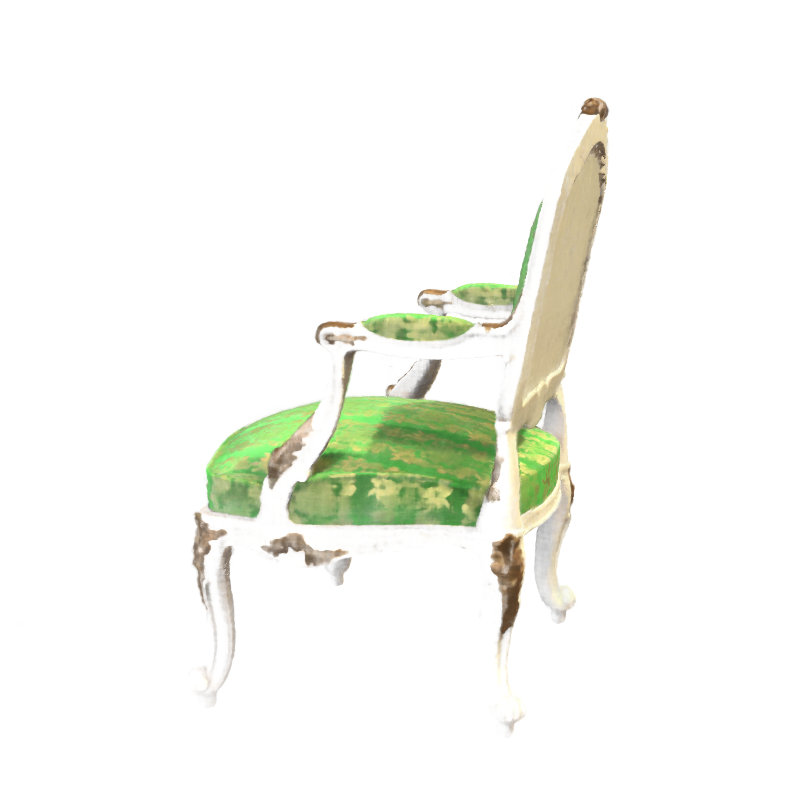}} &
		\raisebox{0.5\imgw}{\rotatebox[origin=c]{-90}{Original}}
		\vspace{-2mm}
		\\
		\raisebox{5mm}{
			\begin{tikzpicture}[x=\palew,y=\palew]
				\node[] (before) at (0,0)
				{\adjustbox{frame}{\includegraphics[height=\palew, angle=-90]{res/our/chair/pale.png}}};
				\node[] (after) at (2,0)
				{\adjustbox{frame}{\includegraphics[height=\palew, angle=-90]{res/our/chair/pale-e0.png}}};
				\draw[->, semithick] (0.5,-1) -- (1.5,-1);
				\draw[->, semithick] (0.5,-2) -- (1.5,-2);
			\end{tikzpicture}
		}
		&
		\adjustbox{}{\includegraphics[width=\imgw]{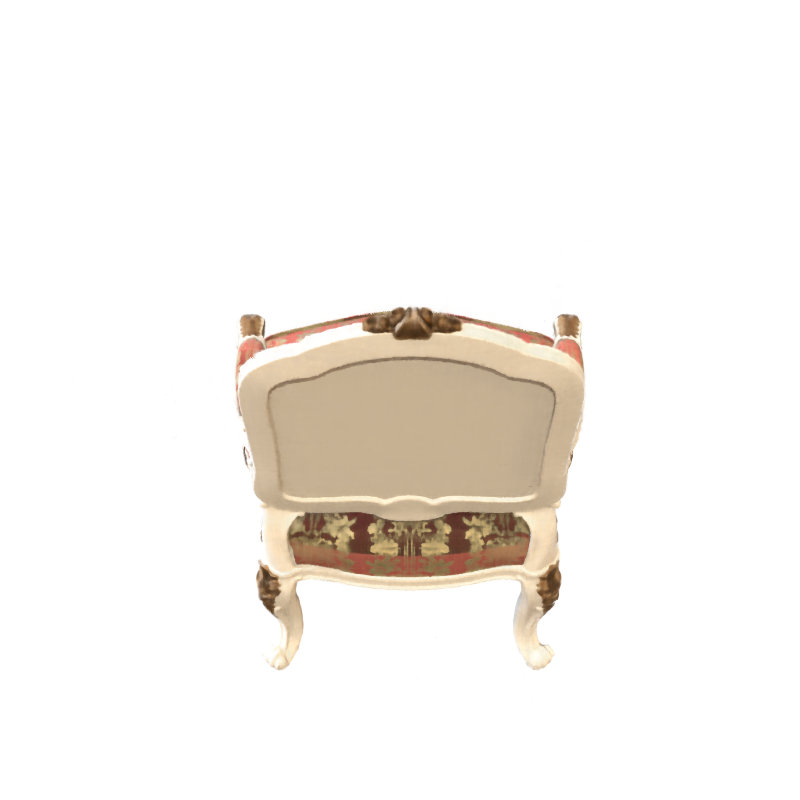}} &
		\adjustbox{}{\includegraphics[width=\imgw]{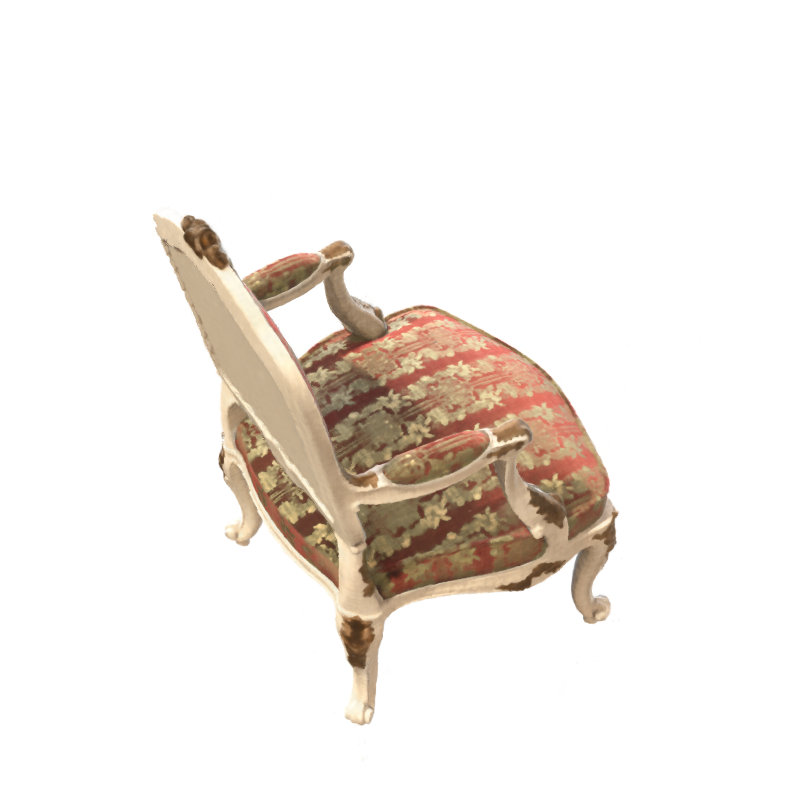}} &
		\adjustbox{}{\includegraphics[width=\imgw]{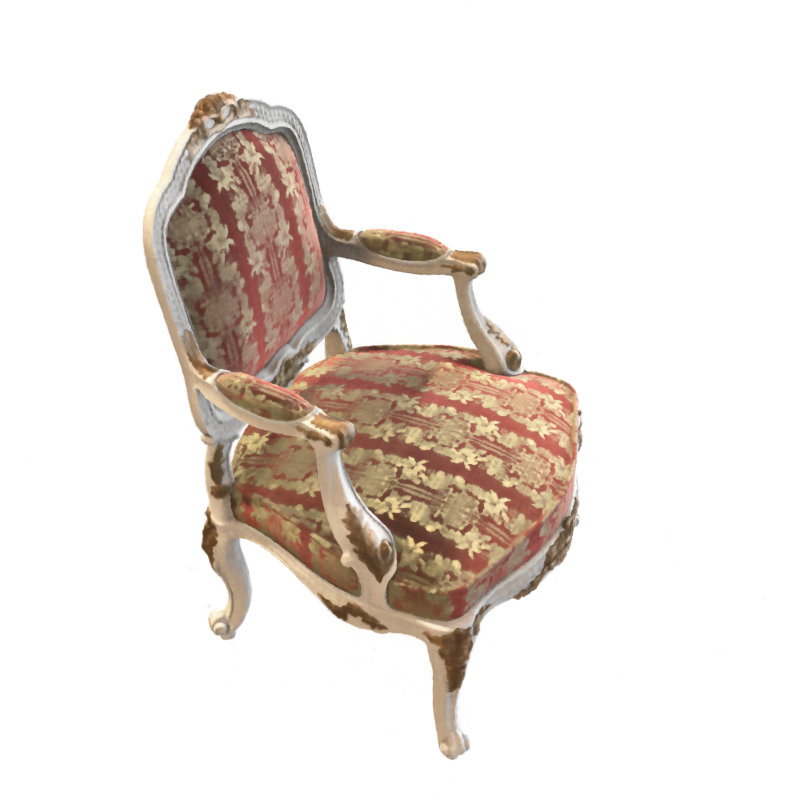}} &
		\adjustbox{}{\includegraphics[width=\imgw]{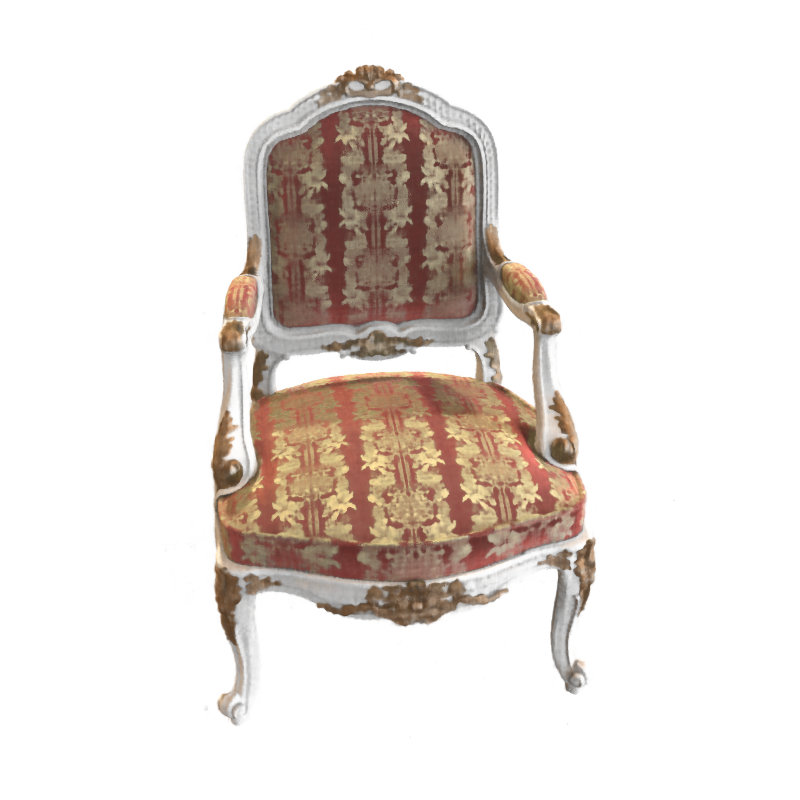}} &
		\adjustbox{}{\includegraphics[width=\imgw]{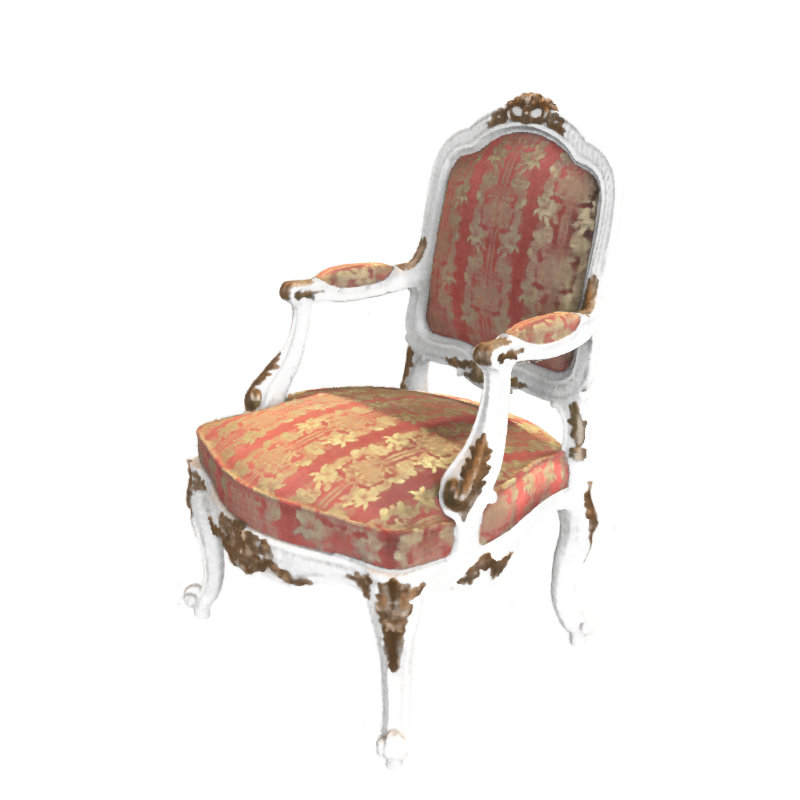}} &
		\adjustbox{}{\includegraphics[width=\imgw]{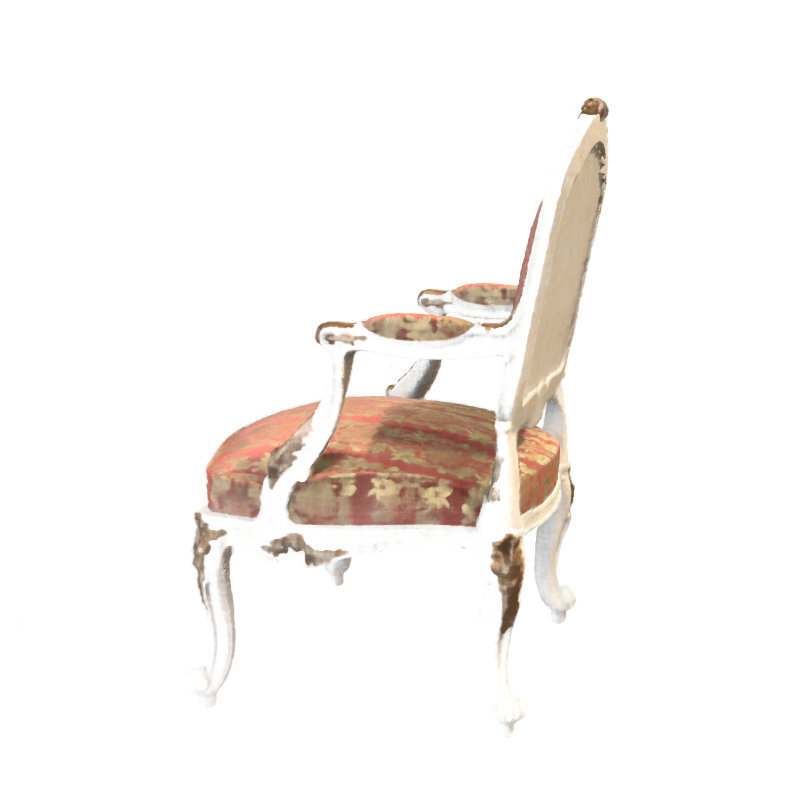}} &
		\raisebox{0.5\imgw}{\rotatebox[origin=c]{-90}{Edited}}
		
		\\
		\hline \vspace{-4pt} 
		\\
		\raisebox{0.375\imgw}{flower} & 
		\adjustbox{}{\includegraphics[width=\imgw]{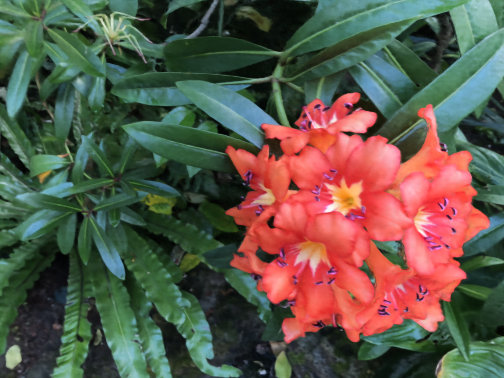}} &
		\adjustbox{}{\includegraphics[width=\imgw]{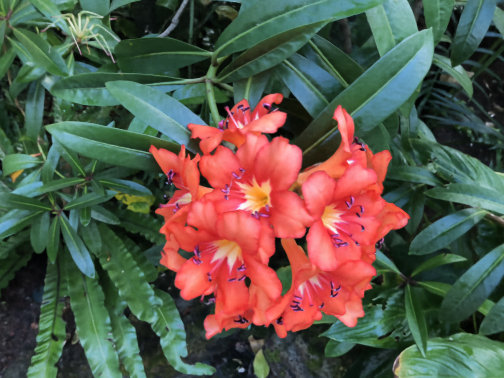}} &
		\adjustbox{}{\includegraphics[width=\imgw]{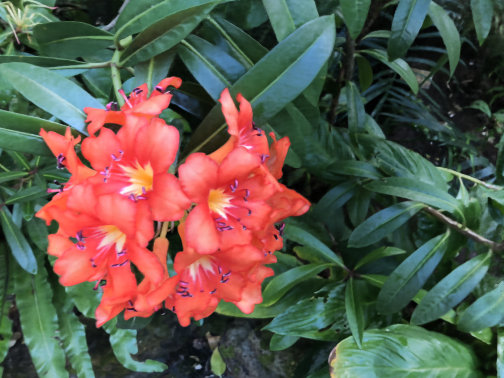}} &
		\adjustbox{}{\includegraphics[width=\imgw]{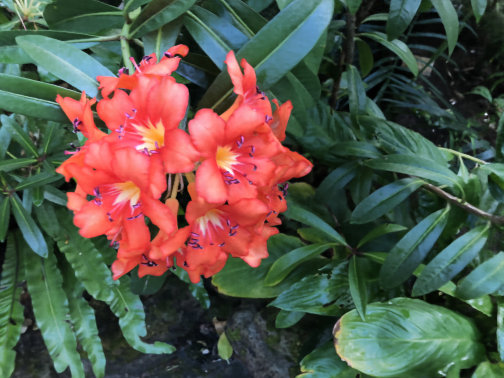}} &
		\adjustbox{}{\includegraphics[width=\imgw]{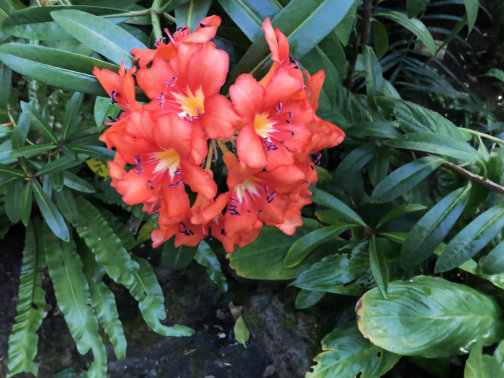}} &
		\adjustbox{}{\includegraphics[width=\imgw]{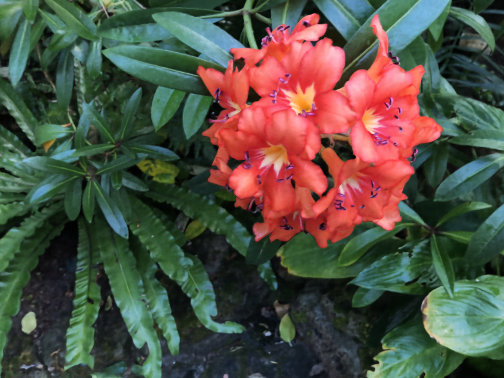}} &
		\raisebox{0.375\imgw}{\rotatebox[origin=c]{-90}{Original}}
		\\
		\specialrule{0pt}{0pt}{0pt}
		\raisebox{-0.8mm}{
			\begin{tikzpicture}[x=\palew,y=\palew]
				\node[] (before) at (0,0)
				{\adjustbox{frame}{\includegraphics[height=\palew, angle=-90]{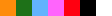}}};
				\node[] (after) at (2,0)
				{\adjustbox{frame}{\includegraphics[height=\palew, angle=-90]{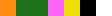}}};
				\draw[->, semithick] (0.5,-1) -- (1.5,-1);
				\draw[->, semithick] (0.5,1) -- (1.5,1);
			\end{tikzpicture}
		}
		&
		\adjustbox{}{\includegraphics[width=\imgw]{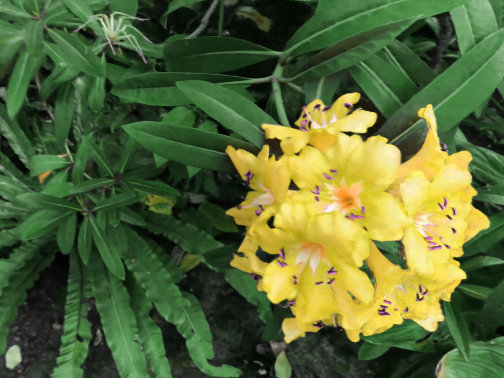}} &
		\adjustbox{}{\includegraphics[width=\imgw]{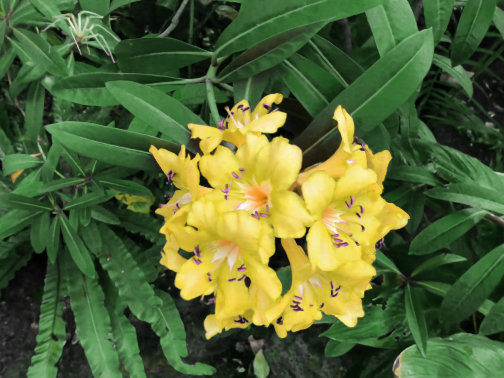}} &
		\adjustbox{}{\includegraphics[width=\imgw]{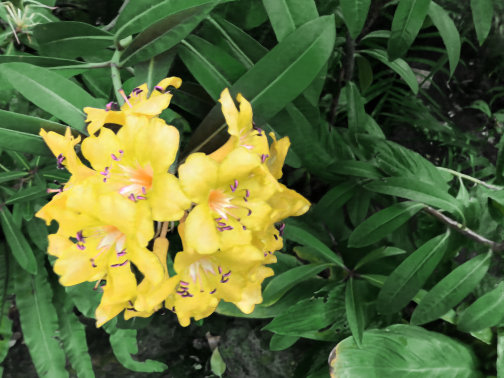}} &
		\adjustbox{}{\includegraphics[width=\imgw]{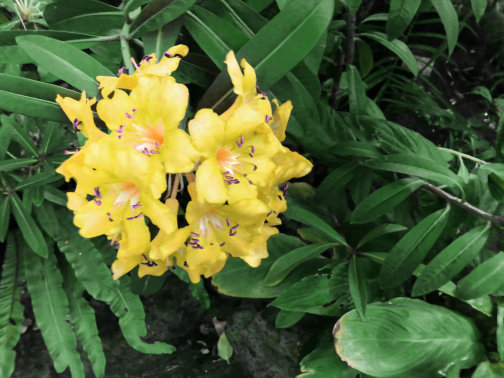}} &
		\adjustbox{}{\includegraphics[width=\imgw]{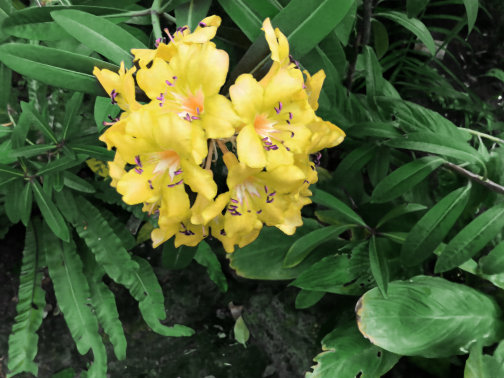}} &
		\adjustbox{}{\includegraphics[width=\imgw]{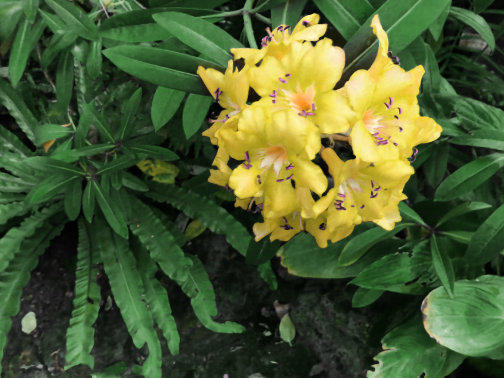}} &
		\raisebox{0.375\imgw}{\rotatebox[origin=c]{-90}{Edited}}

		\\
		\hline \vspace{-4pt} 
		\\
		\raisebox{0.375\imgw}{orchids} & 
		\includegraphics[width=\imgw]{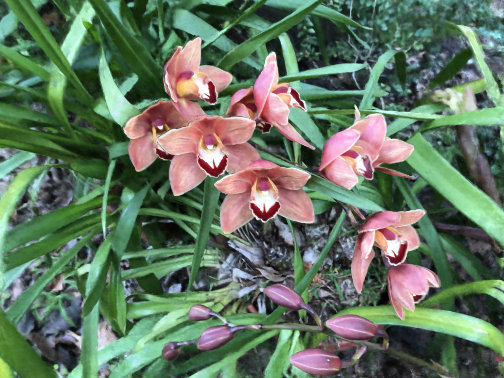} &
		\includegraphics[width=\imgw]{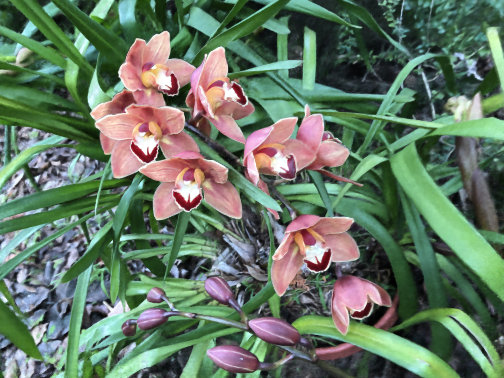} &
		\includegraphics[width=\imgw]{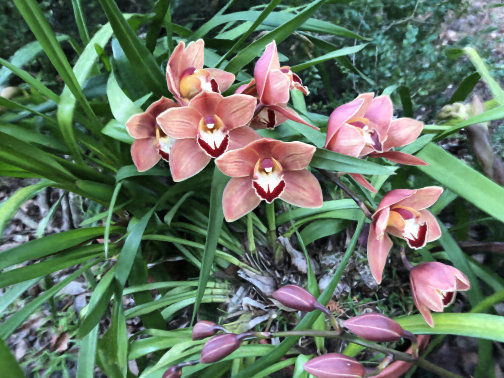} &
		\includegraphics[width=\imgw]{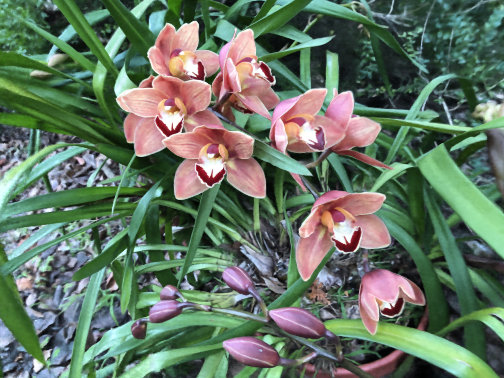} &
		\includegraphics[width=\imgw]{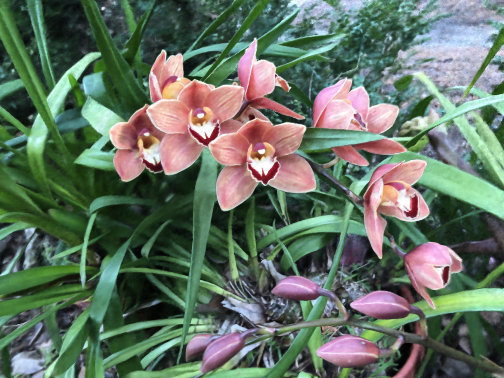} &
		\includegraphics[width=\imgw]{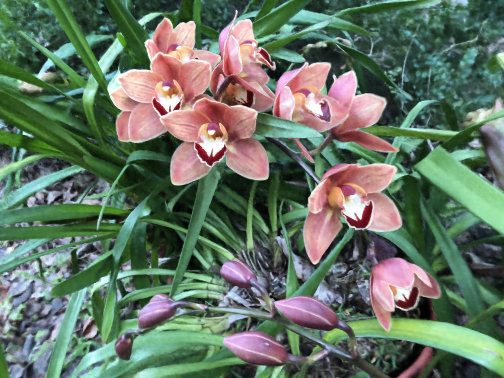} &
		\raisebox{0.375\imgw}{\rotatebox[origin=c]{-90}{Original}}
		\\
		\raisebox{0mm}{
			\begin{tikzpicture}[x=\palew,y=\palew]
				\node[] (before) at (0,0)
				{\adjustbox{frame}{\includegraphics[height=\palew, angle=-90]{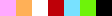}}};
				\node[] (after) at (2,0)
				{\adjustbox{frame}{\includegraphics[height=\palew, angle=-90]{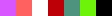}}};
				\draw[->, semithick] (0.5,3) -- (1.5,3);
				\draw[->, semithick] (0.5,2) -- (1.5,2);
				\draw[->, semithick] (0.5,-1) -- (1.5,-1);
			\end{tikzpicture}
		}
		&
		\includegraphics[width=\imgw]{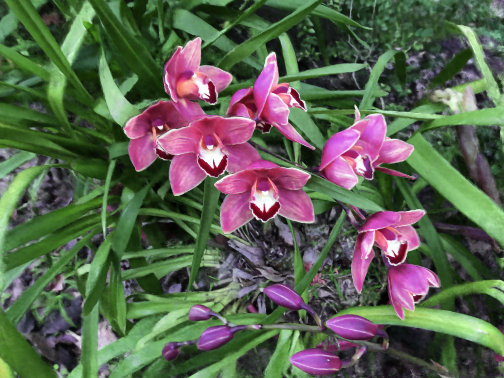} &
		\includegraphics[width=\imgw]{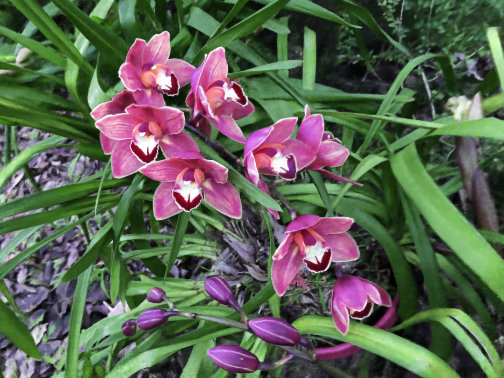} &
		\includegraphics[width=\imgw]{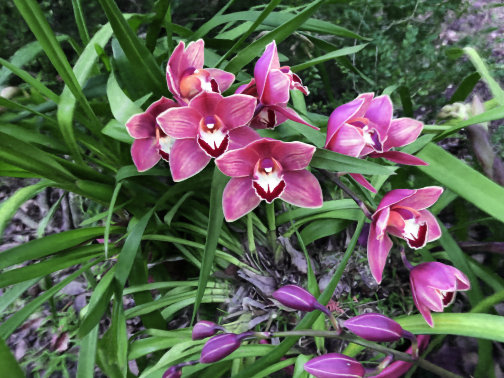} &
		\includegraphics[width=\imgw]{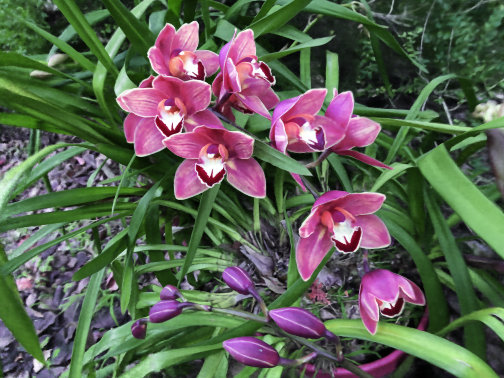} &
		\includegraphics[width=\imgw]{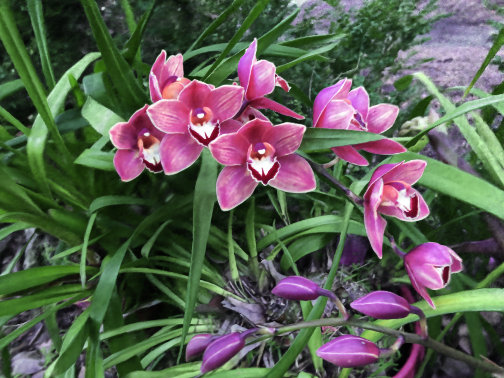} &
		\includegraphics[width=\imgw]{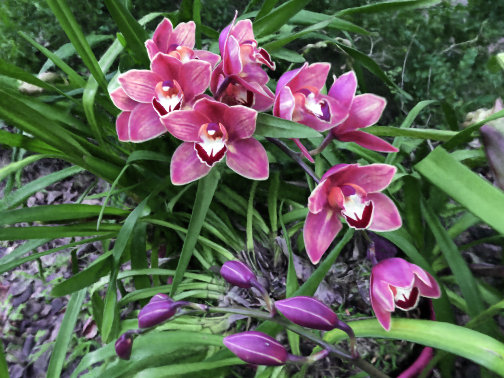} &
		\raisebox{0.375\imgw}{\rotatebox[origin=c]{-90}{Edited}}
		
		\\
		\hline \vspace{-4pt} 
		\\
		\raisebox{0.375\imgw}{vasedeck} & 
		\includegraphics[width=\imgw]{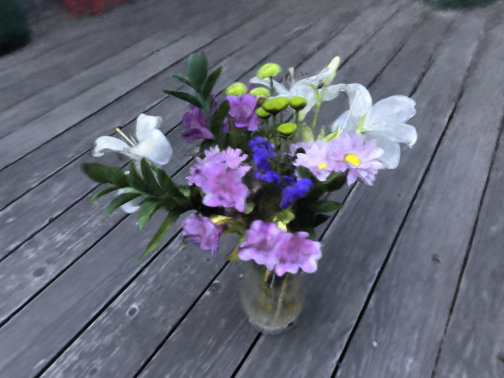} &
		\includegraphics[width=\imgw]{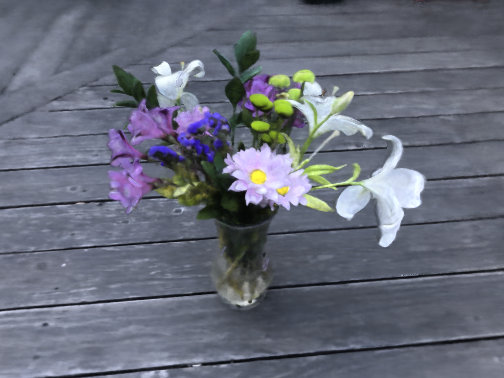} &
		\includegraphics[width=\imgw]{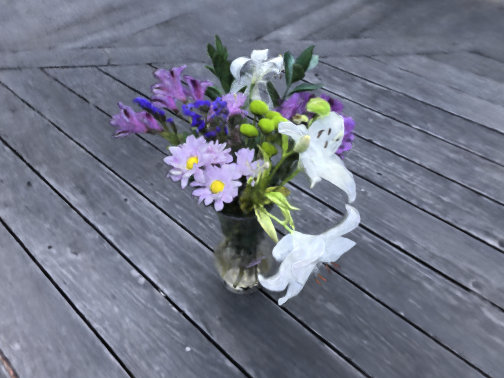} &
		\includegraphics[width=\imgw]{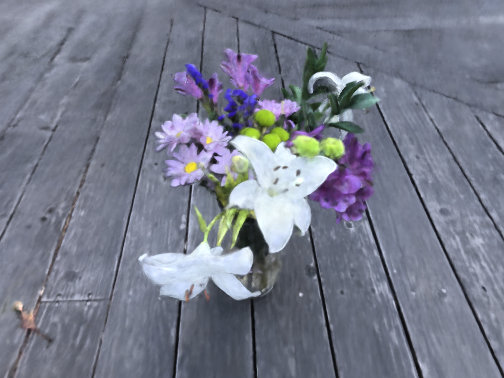} &
		\includegraphics[width=\imgw]{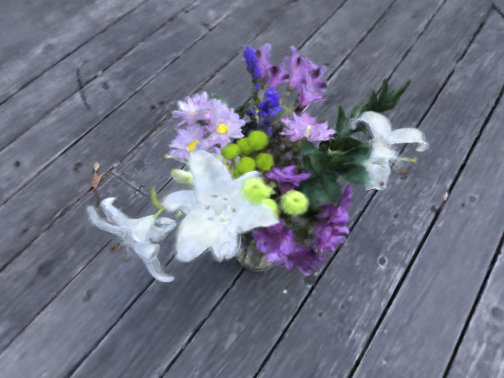} &
		\includegraphics[width=\imgw]{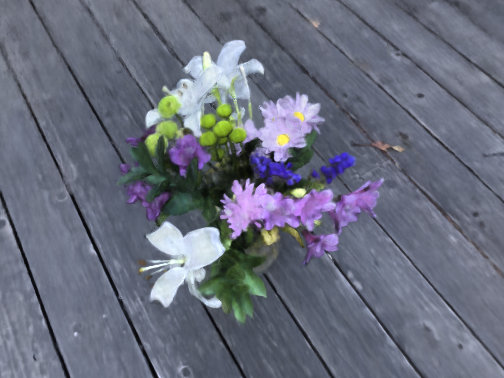} &
		\raisebox{0.375\imgw}{\rotatebox[origin=c]{-90}{Original}}
		\vspace{-1mm} \\
		\raisebox{-1mm}{
			\begin{tikzpicture}[x=\palew,y=\palew]
				\node[] (before) at (0,0)
				{\adjustbox{frame}{\includegraphics[height=\palew, angle=-90]{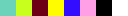}}};
				\node[] (after) at (2,0)
				{\adjustbox{frame}{\includegraphics[height=\palew, angle=-90]{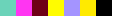}}};
				\draw[->, semithick] (0.5,2.5) -- (1.5,2.5);
				\draw[->, semithick] (0.5,-0.5) -- (1.5,-0.5);
				\draw[->, semithick] (0.5,-1.5) -- (1.5,-1.5);
			\end{tikzpicture}
		}
		&
		\includegraphics[width=\imgw]{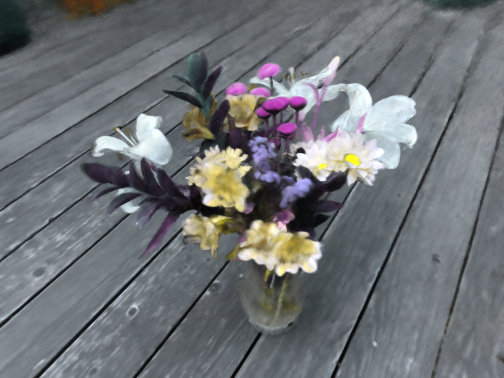} &
		\includegraphics[width=\imgw]{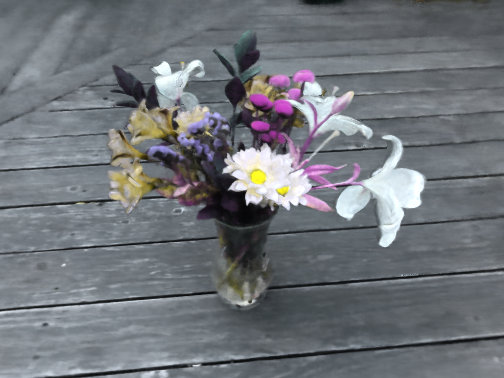} &
		\includegraphics[width=\imgw]{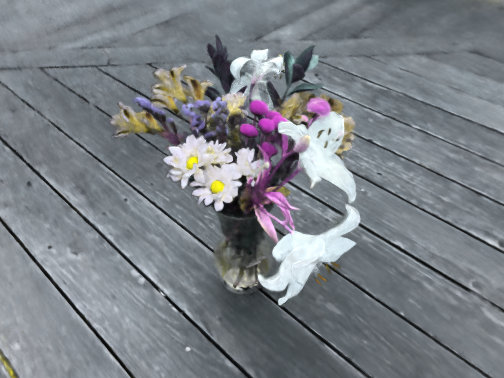} &
		\includegraphics[width=\imgw]{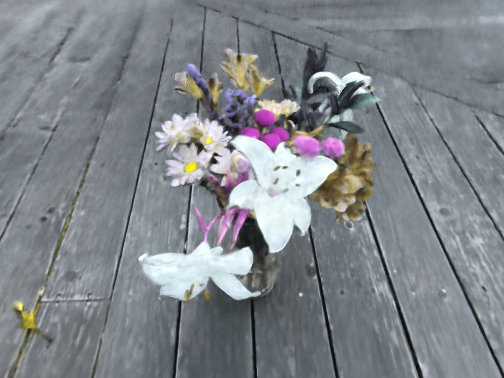} &
		\includegraphics[width=\imgw]{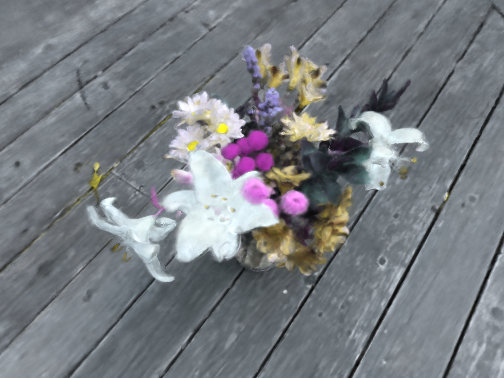} &
		\includegraphics[width=\imgw]{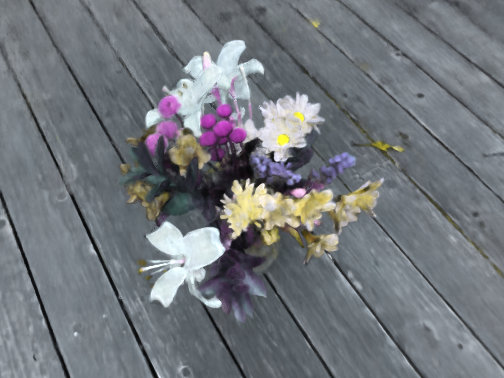} &
		\raisebox{0.375\imgw}{\rotatebox[origin=c]{-90}{Edited}}

		\\
		\hline \vspace{-4pt} 
		\\
		\raisebox{0.333\imgw}{valley} & 
		\includegraphics[width=\imgw]{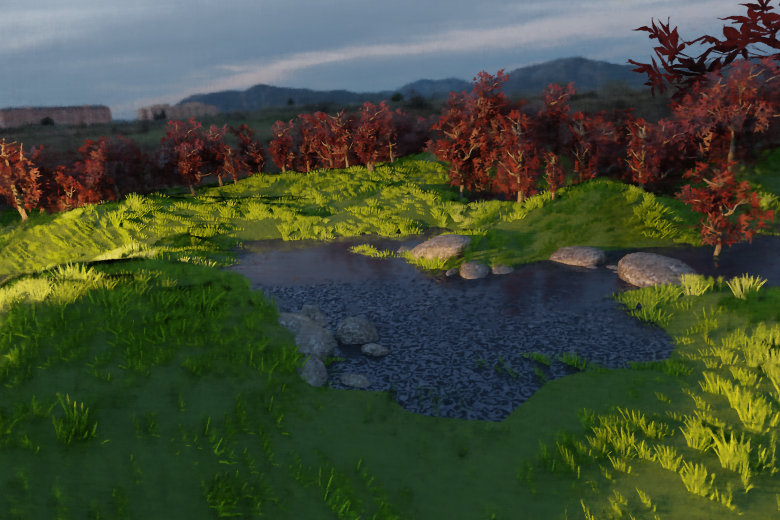} &
		\includegraphics[width=\imgw]{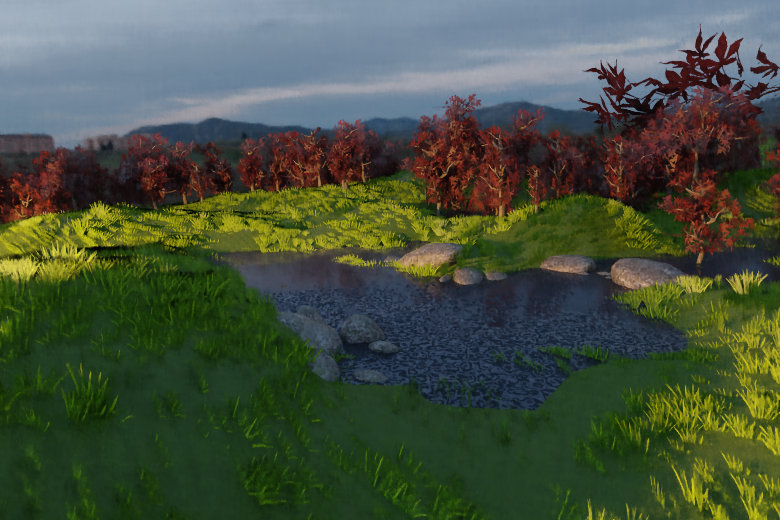} &
		\includegraphics[width=\imgw]{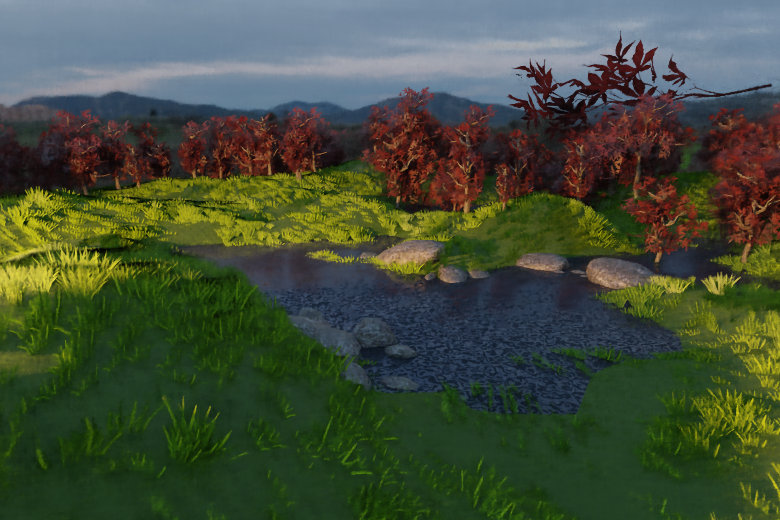} &
		\includegraphics[width=\imgw]{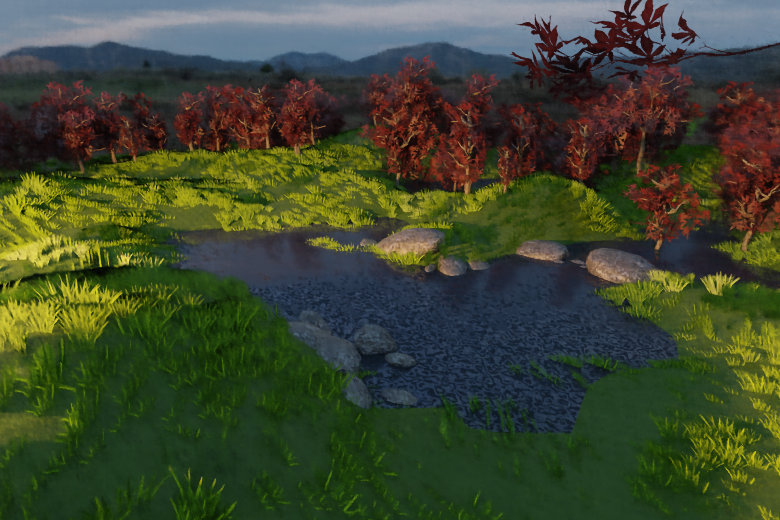} &
		\includegraphics[width=\imgw]{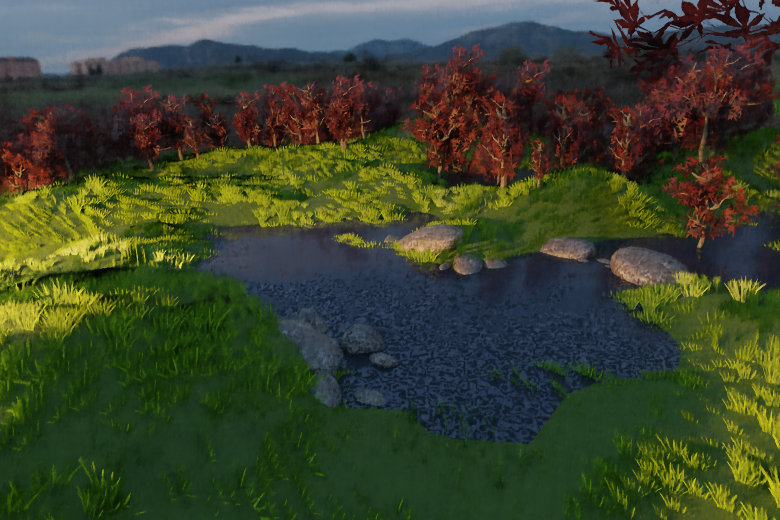} &
		\includegraphics[width=\imgw]{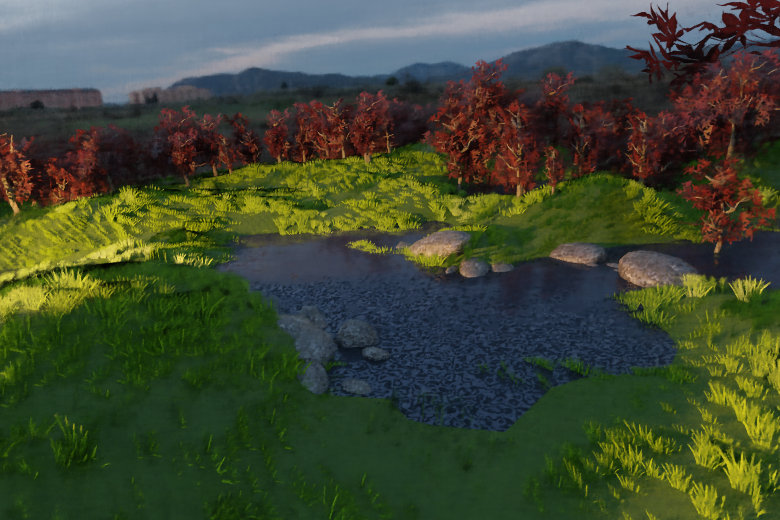} &
		\raisebox{0.333\imgw}{\rotatebox[origin=c]{-90}{Original}}
		\\
		\raisebox{0mm}{
			\begin{tikzpicture}[x=\palew,y=\palew]
				\node[] (before) at (0,0)
				{\adjustbox{frame}{\includegraphics[height=\palew, angle=-90]{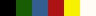}}};
				\node[] (after) at (2,0)
				{\adjustbox{frame}{\includegraphics[height=\palew, angle=-90]{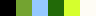}}};
				\draw[->, semithick] (0.5,1.5) -- (1.5,1.5);
				\draw[->, semithick] (0.5,0.5) -- (1.5,0.5);
				\draw[->, semithick] (0.5,-0.5) -- (1.5,-0.5);
				\draw[->, semithick] (0.5,-1.5) -- (1.5,-1.5);
			\end{tikzpicture}
		}
		&
		\includegraphics[width=\imgw]{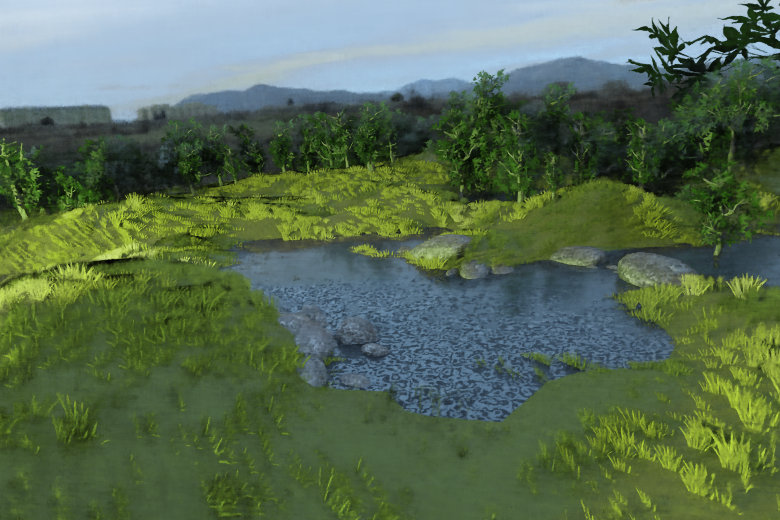} &
		\includegraphics[width=\imgw]{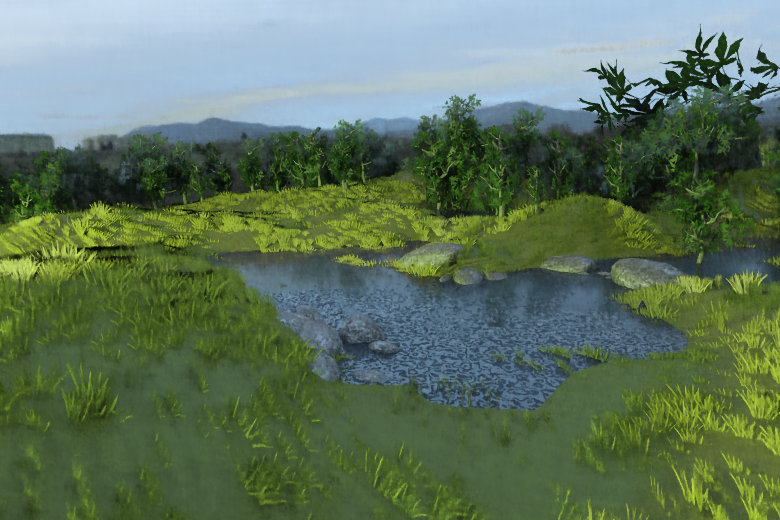} &
		\includegraphics[width=\imgw]{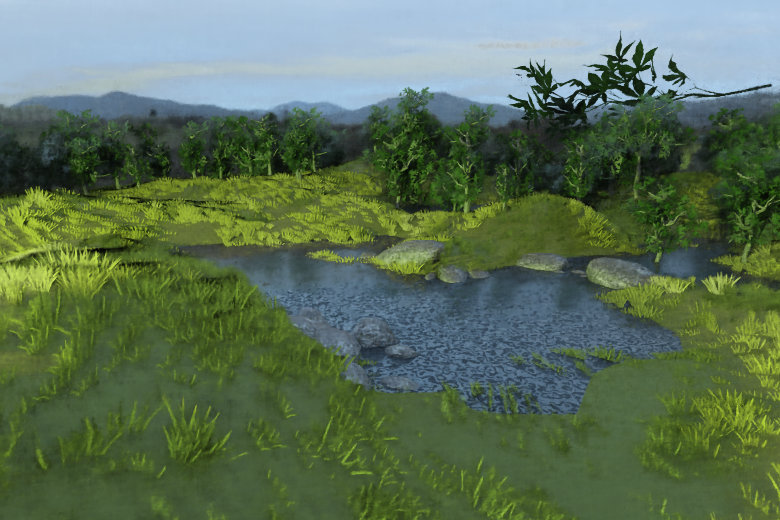} &
		\includegraphics[width=\imgw]{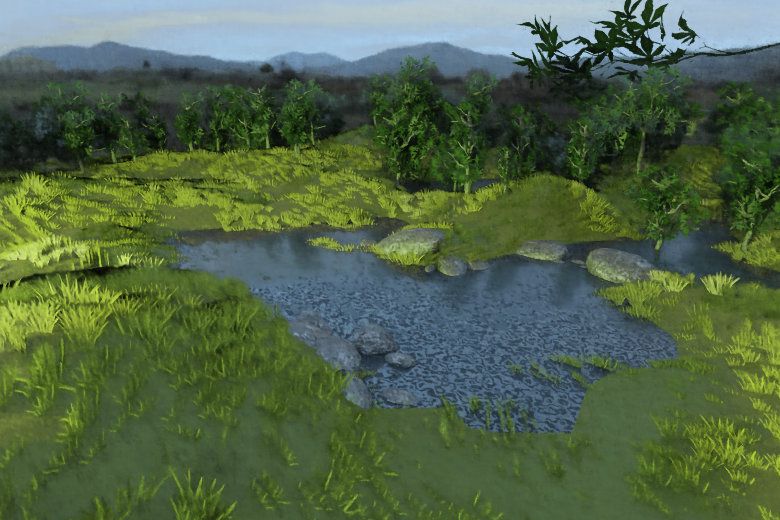} &
		\includegraphics[width=\imgw]{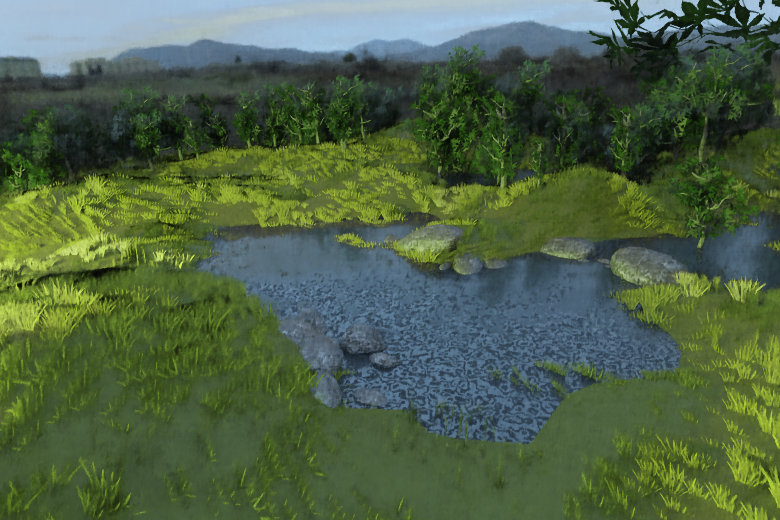} &
		\includegraphics[width=\imgw]{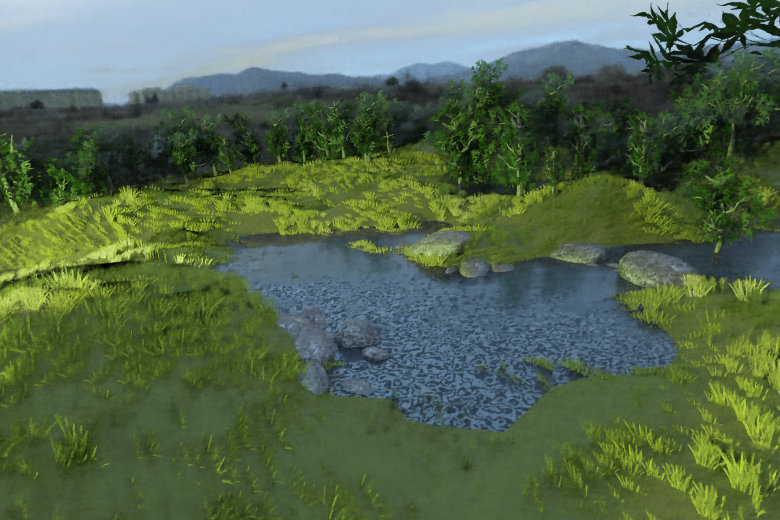} &
		\raisebox{0.333\imgw}{\rotatebox[origin=c]{-90}{Edited 1}}
		\\
		\raisebox{0mm}{
			\begin{tikzpicture}[x=\palew,y=\palew]
				\node[] (before) at (0,0)
				{\adjustbox{frame}{\includegraphics[height=\palew, angle=-90]{res/our/valley/pale.png}}};
				\node[] (after) at (2,0)
				{\adjustbox{frame}{\includegraphics[height=\palew, angle=-90]{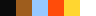}}};
				\draw[->, semithick] (0.5,1.5) -- (1.5,1.5);
				\draw[->, semithick] (0.5,0.5) -- (1.5,0.5);
				\draw[->, semithick] (0.5,-0.5) -- (1.5,-0.5);
				\draw[->, semithick] (0.5,-1.5) -- (1.5,-1.5);
			\end{tikzpicture}
		}
		&
		\includegraphics[width=\imgw]{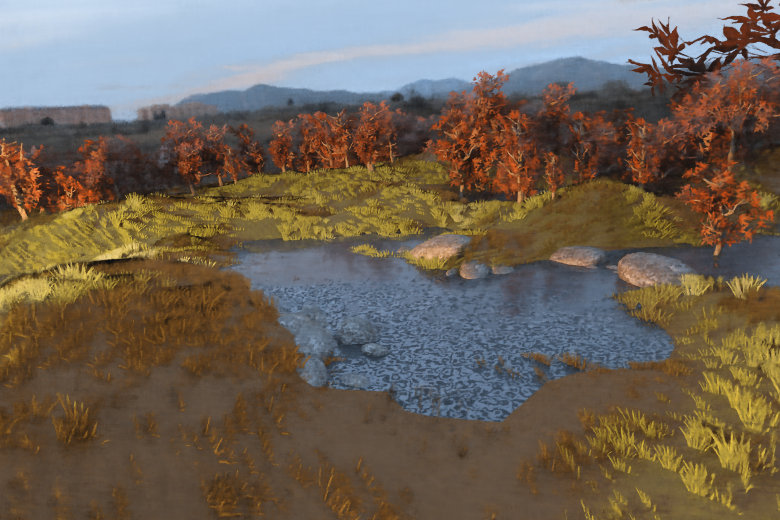} &
		\includegraphics[width=\imgw]{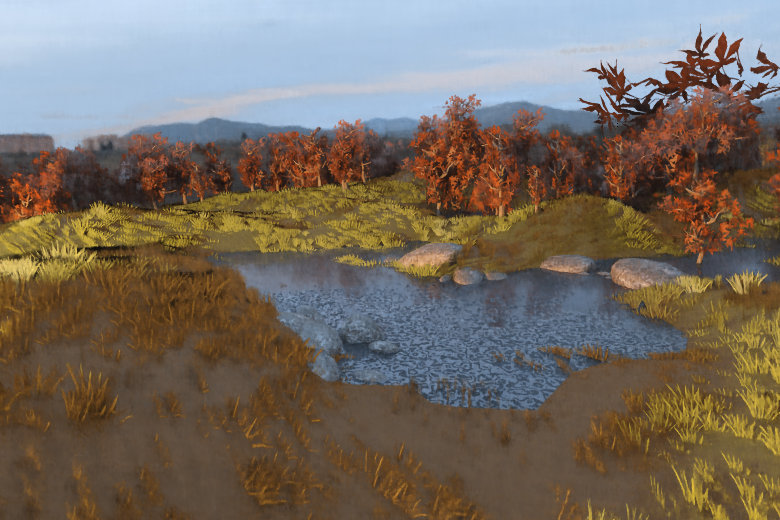} &
		\includegraphics[width=\imgw]{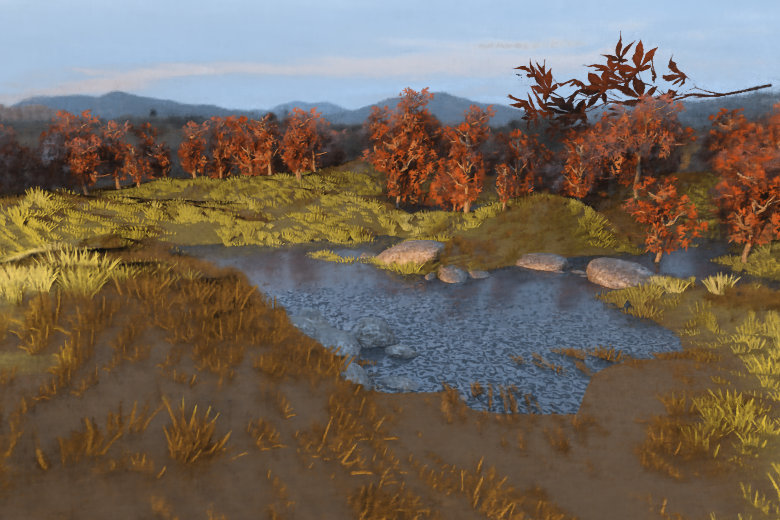} &
		\includegraphics[width=\imgw]{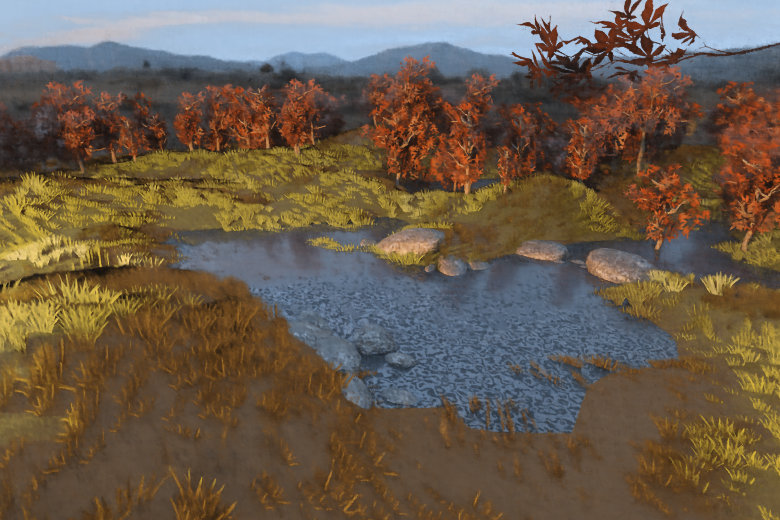} &
		\includegraphics[width=\imgw]{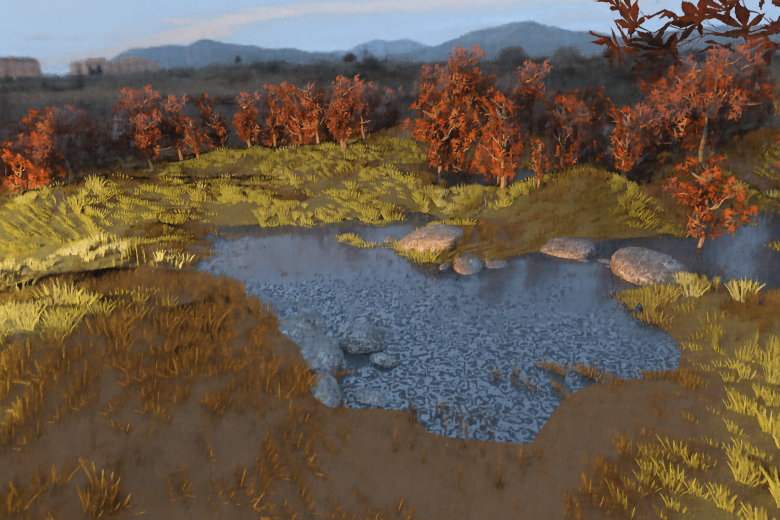} &
		\includegraphics[width=\imgw]{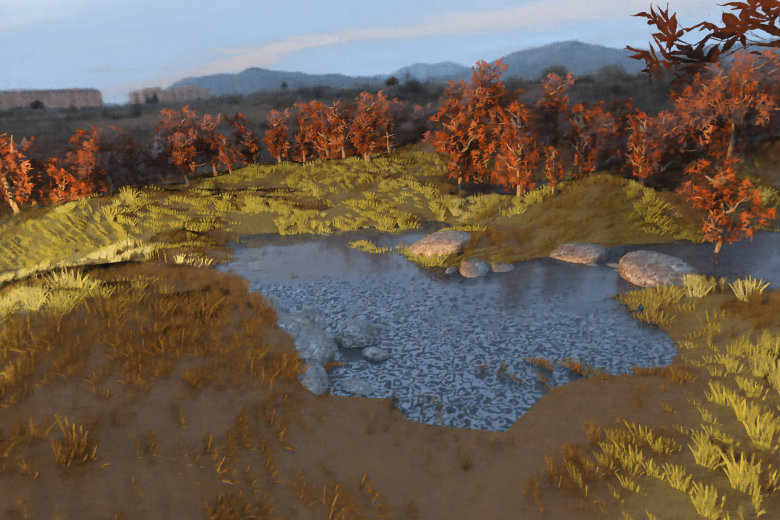} &
		\raisebox{0.333\imgw}{\rotatebox[origin=c]{-90}{Edited 2}}
	\end{tabular}
	
	\caption{
	Color editing results of our method. In each 2 rows, the top one is NeRF's output, the bottom one is editing results of \palenerf.
	}
	\label{fig:real-img}
\end{figure*}

In Figures~\ref{fig:teaser} and \ref{fig:real-img}, we show the recoloring results on several scenes using our \palenerf. For each example, the recolored results on multiple novel views are provided. \palenerf can produce consistent recoloring results across multiple views.
In Table~\ref{tab:real-metric}, we also provide the quantitative reconstruction scores (including PSNR, SSIM, and LPIPS) of \palenerf for all input scenes. The reconstruction score of the vanilla NeRF is also provided. Figure \ref{fig:compare-real} gives visual comparisons between the reconstructed results of our \palenerf and the vanilla NeRF. It could be found that the results quality of \palenerf is comparable to the vanilla NeRF from both aspects of quantitative measures and visual results. Without reducing visual qualities, our method enables the ability of color editing for NeRFs.



\section{Conclusion}
In this paper, we present \palenerf, which unifies the palette-based image decomposition methods and NeRFs to enable color editing of NeRF-represented scenes. Our method is intuitive, efficient, view-consistent, and artifact-free. The users could recolor the scene by adjusting palette colors, and previewing the recolored results from any novel views. 
Moreover, the editing process is decoupled from the network structure, 
which means the backbone of \palenerf and the data preparation step can be replaced with more advanced follow-ups.

Nevertheless, several limitations still exist in our method. For example, we only allow global color editing, hence, if a scene contains two red apples, it is not allowed to only change the color of one apple. Furthermore, geometry editing is not supported. It is worthwhile to investigate ways to edit geometries. 

\clearpage
{\small
\bibliographystyle{cvm}
\bibliography{refs.bib}

\begin{thebibliography}{10}\itemsep=-1pt

\bibitem{aharoni-mack_pigment-based_2017}
E.~Aharoni-Mack, Y.~Shambik, and D.~Lischinski.
\newblock Pigment-based recoloring of watercolor paintings.
\newblock In {\em Proceedings of the {Symposium} on {Non}-{Photorealistic}
  {Animation} and {Rendering}}, pages 1--11, 2017.

\bibitem{aksoy_unmixing-based_2017}
Y.~Aksoy, T.~O. Aydin, A.~Smolić, and M.~Pollefeys.
\newblock Unmixing-based soft color segmentation for image manipulation.
\newblock {\em ACM Transactions on Graphics (TOG)}, 36(2):1--19, 2017.
\newblock Publisher: ACM New York, NY, USA.

\bibitem{cao_mining_2017}
Y.~Cao, A.~B. Chan, and R.~W. Lau.
\newblock Mining probabilistic color palettes for summarizing color use in
  artwork collections.
\newblock In {\em {SIGGRAPH} {Asia} 2017 {Symposium} on {Visualization}}, pages
  1--8, 2017.

\bibitem{chang_palette-based_2015}
H.~Chang, O.~Fried, Y.~Liu, S.~DiVerdi, and A.~Finkelstein.
\newblock Palette-based photo recoloring.
\newblock {\em ACM Trans. Graph.}, 34(4):139--1, 2015.

\bibitem{chao_posterchild_2021}
C.-K.~T. Chao, K.~Singh, and Y.~Gingold.
\newblock {PosterChild}: {Blend}-{Aware} {Artistic} {Posterization}.
\newblock In {\em Computer {Graphics} {Forum}}, volume~40, pages 87--99. Wiley
  Online Library, 2021.
\newblock Issue: 4.

\bibitem{dellaert_survey-nerf_2021}
F.~Dellaert and L.~Yen-Chen.
\newblock Neural {Volume} {Rendering}: {NeRF} {And} {Beyond}.
\newblock {\em arXiv:2101.05204 [cs]}, Jan. 2021.
\newblock arXiv: 2101.05204.

\bibitem{delos_rgb_2019}
B.~Delos, N.~Mellado, D.~Vanderhaeghe, and R.~Cozot.
\newblock {RGB} {Point} {Cloud} {Manipulation} with {Triangular} {Structures}
  for {Artistic} {Image} {Recoloring}.
\newblock {\em arXiv preprint arXiv:1912.04583}, 2019.

\bibitem{dosovitskiy_carla_2017}
A.~Dosovitskiy, G.~Ros, F.~Codevilla, A.~Lopez, and V.~Koltun.
\newblock {CARLA}: {An} open urban driving simulator.
\newblock In {\em Conference on robot learning}, pages 1--16. PMLR, 2017.

\bibitem{du_video_2021}
Z.-J. Du, K.-X. Lei, K.~Xu, J.~Tan, and Y.~Gingold.
\newblock Video recoloring via spatial-temporal geometric palettes.
\newblock {\em ACM Transactions on Graphics (TOG)}, 40(4):1--16, 2021.
\newblock Publisher: ACM New York, NY, USA.

\bibitem{feng_finding_2018}
Z.~Feng, W.~Yuan, C.~Fu, J.~Lei, and M.~Song.
\newblock Finding intrinsic color themes in images with human visual
  perception.
\newblock {\em Neurocomputing}, 273:395--402, 2018.
\newblock Publisher: Elsevier.

\bibitem{floater_mvc_2005}
M.~S. Floater, G.~Kós, and M.~Reimers.
\newblock Mean value coordinates in {3D}.
\newblock {\em Computer Aided Geometric Design}, 22(7):623--631, 2005.
\newblock Publisher: Elsevier.

\bibitem{garbin_fastnerf_2021}
S.~J. Garbin, M.~Kowalski, M.~Johnson, J.~Shotton, and J.~Valentin.
\newblock {FastNeRF}: {High}-{Fidelity} {Neural} {Rendering} at {200FPS}.
\newblock pages 14346--14355, 2021.

\bibitem{grogan_image_2020}
M.~Grogan and A.~Smolic.
\newblock Image {Decomposition} using {Geometric} {Region} {Colour} {Unmixing}.
\newblock In {\em European {Conference} on {Visual} {Media} {Production}},
  pages 1--10, 2020.

\bibitem{guo_osf_2020}
M.~Guo, A.~Fathi, J.~Wu, and T.~Funkhouser.
\newblock Object-centric neural scene rendering.
\newblock {\em arXiv preprint arXiv:2012.08503}, Dec. 2020.

\bibitem{ju_mvc_2005}
T.~Ju, S.~Schaefer, and J.~Warren.
\newblock Mean value coordinates for closed triangular meshes.
\newblock In {\em {ACM} {Siggraph} 2005 {Papers}}, pages 561--566. 2005.

\bibitem{kim_automatic_2020}
S.~Kim and S.~Choi.
\newblock Automatic {Color} {Scheme} {Extraction} from {Movies}.
\newblock In {\em Proceedings of the 2020 {International} {Conference} on
  {Multimedia} {Retrieval}}, pages 154--163, 2020.

\bibitem{lin_layerbuilder_2017}
S.~Lin, M.~Fisher, A.~Dai, and P.~Hanrahan.
\newblock {LayerBuilder}: {Layer} decomposition for interactive image and video
  color editing.
\newblock {\em arXiv preprint arXiv:1701.03754}, 2017.

\bibitem{lin_modeling_2013}
S.~Lin and P.~Hanrahan.
\newblock Modeling how people extract color themes from images.
\newblock In {\em Proceedings of the {SIGCHI} {Conference} on {Human} {Factors}
  in {Computing} {Systems}}, pages 3101--3110, 2013.

\bibitem{liu_nsvf_2021}
L.~Liu, J.~Gu, K.~Z. Lin, T.-S. Chua, and C.~Theobalt.
\newblock Neural {Sparse} {Voxel} {Fields}.
\newblock In {\em {NeurIPS}}, Jan. 2021.
\newblock arXiv: 2007.11571.

\bibitem{liu_editnerf_2021}
S.~Liu, X.~Zhang, Z.~Zhang, R.~Zhang, J.-Y. Zhu, and B.~Russell.
\newblock Editing conditional radiance fields.
\newblock In {\em Proceedings of the {IEEE}/{CVF} {International} {Conference}
  on {Computer} {Vision}}, pages 5773--5783, 2021.

\bibitem{martin-nerf-inthewild_2021}
R.~Martin-Brualla, N.~Radwan, M.~S.~M. Sajjadi, J.~T. Barron, A.~Dosovitskiy,
  and D.~Duckworth.
\newblock {NeRF} in the {Wild}: {Neural} {Radiance} {Fields} for
  {Unconstrained} {Photo} {Collections}.
\newblock In {\em Proceedings of the {IEEE}/{CVF} {Conference} on {Computer}
  {Vision} and {Pattern} {Recognition}}, Jan. 2021.
\newblock arXiv: 2008.02268.

\bibitem{mildenhall_nerf_2020}
B.~Mildenhall, P.~P. Srinivasan, M.~Tancik, J.~T. Barron, R.~Ramamoorthi, and
  R.~Ng.
\newblock Nerf: {Representing} scenes as neural radiance fields for view
  synthesis.
\newblock In {\em European conference on computer vision}, pages 405--421.
  Springer, Mar. 2020.

\bibitem{muller_instantngp_2022}
T.~Müller, A.~Evans, C.~Schied, and A.~Keller.
\newblock Instant {Neural} {Graphics} {Primitives} with a {Multiresolution}
  {Hash} {Encoding}.
\newblock {\em arXiv:2201.05989 [cs]}, Jan. 2022.
\newblock arXiv: 2201.05989.

\bibitem{nguyen_group-theme_2017}
R.~M. Nguyen, B.~Price, S.~Cohen, and M.~S. Brown.
\newblock Group-{Theme} {Recoloring} for {Multi}-{Image} {Color} {Consistency}.
\newblock In {\em Computer {Graphics} {Forum}}, volume~36, pages 83--92. Wiley
  Online Library, 2017.
\newblock Issue: 7.

\bibitem{niemeyer_giraffe_2021}
M.~Niemeyer and A.~Geiger.
\newblock {GIRAFFE}: {Representing} {Scenes} as {Compositional} {Generative}
  {Neural} {Feature} {Fields}.
\newblock In {\em {CVPR}}, page~12, 2021.

\bibitem{odonovan_color_2011}
P.~O'Donovan, A.~Agarwala, and A.~Hertzmann.
\newblock Color compatibility from large datasets.
\newblock In {\em {ACM} {SIGGRAPH} 2011 papers}, pages 1--12. 2011.

\bibitem{ost_neural-scene-graph_2021}
J.~Ost, F.~Mannan, N.~Thuerey, J.~Knodt, and F.~Heide.
\newblock Neural scene graphs for dynamic scenes.
\newblock In {\em Proceedings of the {IEEE}/{CVF} {Conference} on {Computer}
  {Vision} and {Pattern} {Recognition}}, pages 2856--2865, 2021.

\bibitem{park_nerfies_2021}
K.~Park, U.~Sinha, J.~T. Barron, S.~Bouaziz, D.~B. Goldman, S.~M. Seitz, and
  R.~Martin-Brualla.
\newblock Nerfies: {Deformable} {Neural} {Radiance} {Fields}.
\newblock In {\em Proceedings of the {IEEE}/{CVF} {International} {Conference}
  on {Computer} {Vision}}, pages 5865--5874, 2021.

\bibitem{park_hypernerf_2021}
K.~Park, U.~Sinha, P.~Hedman, J.~T. Barron, S.~Bouaziz, D.~B. Goldman,
  R.~Martin-Brualla, and S.~M. Seitz.
\newblock {HyperNeRF}: {A} {Higher}-{Dimensional} {Representation} for
  {Topologically} {Varying} {Neural} {Radiance} {Fields}.
\newblock {\em arXiv:2106.13228 [cs]}, June 2021.
\newblock arXiv: 2106.13228.

\bibitem{pbrt-book-2016}
M.~Pharr, W.~Jakob, and G.~Humphreys.
\newblock {\em Physically based rendering: From theory to implementation}.
\newblock Morgan Kaufmann, 2016.

\bibitem{pumarola_d-nerf_2021}
A.~Pumarola, E.~Corona, G.~Pons-Moll, and F.~Moreno-Noguer.
\newblock D-nerf: {Neural} radiance fields for dynamic scenes.
\newblock In {\em Proceedings of the {IEEE}/{CVF} {Conference} on {Computer}
  {Vision} and {Pattern} {Recognition}}, pages 10318--10327, 2021.

\bibitem{radford_clip_2021}
A.~Radford, J.~W. Kim, C.~Hallacy, A.~Ramesh, G.~Goh, S.~Agarwal, G.~Sastry,
  A.~Askell, P.~Mishkin, and J.~Clark.
\newblock Learning transferable visual models from natural language
  supervision.
\newblock In {\em International {Conference} on {Machine} {Learning}}, pages
  8748--8763. PMLR, 2021.

\bibitem{reiser_kilonerf_2021}
C.~Reiser, S.~Peng, Y.~Liao, and A.~Geiger.
\newblock {KiloNeRF}: {Speeding} up {Neural} {Radiance} {Fields} with
  {Thousands} of {Tiny} {MLPs}.
\newblock In {\em {ICCV} 2021}, 2021.

\bibitem{Tan:2018:PPB}
J.~Tan, S.~{DiVerdi}, J.~Lu, and Y.~Gingold.
\newblock Pigmento: Pigment-based image analysis and editing.
\newblock {\em Transactions on Visualization and Computer Graphics (TVCG)},
  25(9), 2019.

\bibitem{tan_decomposing_2015}
J.~Tan, M.~Dvorožňák, D.~Sykora, and Y.~Gingold.
\newblock Decomposing time-lapse paintings into layers.
\newblock {\em ACM Transactions on Graphics (TOG)}, 34(4):1--10, 2015.
\newblock Publisher: ACM New York, NY, USA.

\bibitem{tan_palette-based_2018}
J.~Tan, J.~Echevarria, and Y.~Gingold.
\newblock Palette-based image decomposition, harmonization, and color transfer.
\newblock Apr. 2018.

\bibitem{tan_efficient_2019}
J.~Tan, J.~Echevarria, and Y.~Gingold.
\newblock Efficient palette-based decomposition and recoloring of images via
  {RGBXY}-space geometry.
\newblock {\em ACM Transactions on Graphics}, 37(6):1--10, Jan. 2019.

\bibitem{tan_decomposing_2017}
J.~Tan, J.-M. Lien, and Y.~Gingold.
\newblock Decomposing {Images} into {Layers} via {RGB}-{Space} {Geometry}.
\newblock {\em ACM Transactions on Graphics}, 36(1):1--14, Feb. 2017.

\bibitem{tewari_survey-neu-render_2022}
A.~Tewari, J.~Thies, B.~Mildenhall, P.~Srinivasan, E.~Tretschk, Y.~Wang,
  C.~Lassner, V.~Sitzmann, R.~Martin-Brualla, S.~Lombardi, T.~Simon,
  C.~Theobalt, M.~Niessner, J.~T. Barron, G.~Wetzstein, M.~Zollhoefer, and
  V.~Golyanik.
\newblock Advances in {Neural} {Rendering}.
\newblock {\em arXiv:2111.05849 [cs]}, Mar. 2022.
\newblock arXiv: 2111.05849.

\bibitem{tojo_poster-nerf_2022}
K.~Tojo and N.~Umetani.
\newblock Recolorable {Posterization} of {Volumetric} {Radiance} {Fields}
  {Using} {Visibility}-{Weighted} {Palette} {Extraction}.
\newblock In {\em Computer {Graphics} {Forum}}, volume~41, pages 149--160.
  Wiley Online Library, 2022.
\newblock Issue: 4.

\bibitem{tretschk_non-rigid_2021}
E.~Tretschk, A.~Tewari, V.~Golyanik, M.~Zollhofer, C.~Lassner, and C.~Theobalt.
\newblock Non-{Rigid} {Neural} {Radiance} {Fields}: {Reconstruction} and
  {Novel} {View} {Synthesis} of a {Dynamic} {Scene} {From} {Monocular} {Video}.
\newblock In {\em Proceedings of the {IEEE}/{CVF} {International} {Conference}
  on {Computer} {Vision}}, pages 12959--12970, 2021.

\bibitem{wang_clip-nerf_2022}
C.~Wang, M.~Chai, M.~He, D.~Chen, and J.~Liao.
\newblock {CLIP}-{NeRF}: {Text}-and-{Image} {Driven} {Manipulation} of {Neural}
  {Radiance} {Fields}.
\newblock In {\em Proceedings of the {IEEE}/{CVF} conference on computer vision
  and pattern recognition}. arXiv, Mar. 2022.
\newblock arXiv:2112.05139 [cs] type: article.

\bibitem{wang_mvc_2019}
Y.~Wang, Y.~Liu, and K.~Xu.
\newblock An {Improved} {Geometric} {Approach} for {Palette}-based {Image}
  {Decomposition} and {Recoloring}.
\newblock In {\em Computer {Graphics} {Forum}}, volume~38, pages 11--22. Wiley
  Online Library, 2019.
\newblock Issue: 7.

\bibitem{yang_compose_2021}
B.~Yang, Y.~Zhang, Y.~Xu, Y.~Li, H.~Zhou, H.~Bao, G.~Zhang, and Z.~Cui.
\newblock Learning object-compositional neural radiance field for editable
  scene rendering.
\newblock In {\em Proceedings of the {IEEE}/{CVF} {International} {Conference}
  on {Computer} {Vision}}, pages 13779--13788, 2021.

\bibitem{yu_plenoctrees_2021}
A.~Yu, R.~Li, M.~Tancik, H.~Li, R.~Ng, and A.~Kanazawa.
\newblock Plenoctrees for real-time rendering of neural radiance fields.
\newblock In {\em Proceedings of the {IEEE}/{CVF} {International} {Conference}
  on {Computer} {Vision}}, pages 5752--5761, 2021.

\bibitem{yu_compose_2022}
H.-X. Yu, L.~J. Guibas, and J.~Wu.
\newblock Unsupervised discovery of object radiance fields.
\newblock In {\em International {Conference} on {Learning} {Representations}},
  2022.

\bibitem{zhang_nerf++_2020}
K.~Zhang, G.~Riegler, N.~Snavely, and V.~Koltun.
\newblock Nerf++: {Analyzing} and improving neural radiance fields.
\newblock {\em arXiv preprint arXiv:2010.07492}, 2020.

\bibitem{zhang_palette-based_2017}
Q.~Zhang, C.~Xiao, H.~Sun, and F.~Tang.
\newblock Palette-based image recoloring using color decomposition
  optimization.
\newblock {\em IEEE Transactions on Image Processing}, 26(4):1952--1964, 2017.
\newblock Publisher: IEEE.

\end{thebibliography}
}

\end{document}